%% file: egpaper_for_review.tex
\documentclass[10pt,twocolumn,letterpaper]{article}

 \usepackage{cvpr}
\usepackage[T1]{fontenc}
\usepackage{newtxtext,newtxmath}
\usepackage{epsfig}
\usepackage{graphicx}
\graphicspath{ {./figures/} }

\usepackage{makecell} 
\usepackage{color}

\usepackage{booktabs}
\usepackage{multirow}
\PassOptionsToPackage{table}{xcolor}
\usepackage[pagebackref=false,breaklinks=true,colorlinks,bookmarks=false]{hyperref}

\usepackage{cleveref}

\input{utils/include}

\input{utils/general_utils}

\input{utils/math_utils}

\begin{document}

%%%%%%%%% TITLE
\title{Foundation Models for Face Presentation Attack Detection: A Unified Linear-Probing Benchmark}
% \title{Size matters: Linear Probing Vision Foundation Models for Face PAD}
%\title{Beyond Fine-Tuning: Linear Probing Vision Foundation Models for Face PAD}

% \author{
% Peter Lorenz, Anjith George, Sebastien Marcel\\
% Idiap Research Institute\\
% Rue Marconi 19, Martigny, Switzerland.\\
% </span>,
% {
% \tt\small \{peter.lorenz, anjith.george, sebastien.marcel\}@idiap.ch}
% }

\author{
Peter Lorenz$^{1}$ \quad
Anjith George$^{1}$ \quad
S\'ebastien Marcel$^{1,2}$\\[0.5em]
$^1$Idiap Research Institute, Martigny, Switzerland\\
$^2$University of Lausanne (UNIL), Lausanne, Switzerland\\[0.5em]
{\tt\small \{peter.lorenz, anjith.george, sebastien.marcel\}@idiap.ch}
}

\maketitle
\thispagestyle{empty}

%%%%%%%%% ABSTRACT
\input{content/abstract}

%%%%%%%%% CONTENT
\input{content/content}

%%%%%%%%% REFERENCE
% \input{output.bbl}
% \FloatBarrier
% {\small
% \bibliographystyle{ieee}
% \bibliography{egbib,foundpad}
% }

%%%%%%%%% APPENDIX
\clearpage
\input{content/supplementary}

\end{document}

%% file: utils/include.tex
\usepackage{microtype}
\usepackage{graphicx}
\usepackage{booktabs} % for professional tables

\usepackage{pifont}
\usepackage{blindtext}
\usepackage{lipsum}
\usepackage{threeparttable}

\usepackage{times}
\usepackage{multirow}
\usepackage{url}

\usepackage{enumitem}
\usepackage{xspace}
\usepackage{xfrac}
\usepackage{comment}
\usepackage{tikz}
\usetikzlibrary{arrows.meta, positioning, shapes.geometric, fit, backgrounds, calc}
\newcommand{\iconfrozen}{\raisebox{-0.12ex}{\includegraphics[height=1.25em]{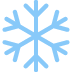}}}
\newcommand{\icontrainable}{\raisebox{-0.12ex}{\includegraphics[height=1.25em]{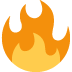}}}
\newcommand{\iconface}{\includegraphics[height=1.15cm]{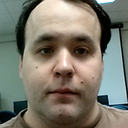}}
\usepackage{etoolbox}

\usepackage{placeins}
\definecolor{copied}{rgb}{0.858, 0.188, 0.478}

\usepackage{pifont}
\usepackage{adjustbox}
\usepackage{threeparttable}
\usepackage{multirow}

\DeclareMathAlphabet\mathbfcal{OMS}{cmsy}{b}{n}
\usepackage{color, colortbl}
\definecolor{Gray}{gray}{0.93}
\definecolor{Orange}{rgb}{1,0.5,0}
\definecolor{DGray}{gray}{0.83}
\definecolor{LightCyan}{rgb}{0.88,1,1}

\usepackage{newfloat}
\DeclareFloatingEnvironment[name={Supplementary Figure}]{suppfigure}
\usepackage{sidecap}

\usepackage{arydshln}

\usepackage{float}

%% file: utils/general_utils.tex
\usepackage{pifont}

\newcommand{\bestcell}[1]{\textbf{#1}}
\definecolor{Orange}{rgb}{1,0.5,0}
\definecolor{DGray}{gray}{0.83}
\definecolor{LightCyan}{rgb}{0.88,1,1}
\newcommand{\htablegrayrow}[1]{%
  \rowcolor{Gray}%
  #1%
}

\usepackage[most]{tcolorbox}

%% file: utils/math_utils.tex
\usepackage{amsmath,amsfonts,bm}

\def\eqref#1{(\ref{#1})}
\def\1{\bm{1}}

\DeclareMathAlphabet{\mathsfit}{\encodingdefault}{\sfdefault}{m}{sl}
\SetMathAlphabet{\mathsfit}{bold}{\encodingdefault}{\sfdefault}{bx}{n}

%% file: content/abstract.tex
\begin{abstract}
Face presentation attack detection (PAD) remains challenging under cross-dataset evaluation, where domain shift degrades models trained on a single dataset. The scarcity of large-scale labeled data motivates adapting pretrained vision models rather than training task-specific architectures from scratch, raising a fundamental question: do general-purpose vision foundation models encode PAD-relevant information accessible with minimal task-specific training? To investigate, we systematically evaluate 24 frozen encoders, including self-supervised Vision Transformers, vision–language encoders, and supervised CNNs, using a unified linear-probing protocol on the MCIO benchmark (MSU-MFSD, CASIA-FASD, Replay-Attack, OULU-NPU). 
The backbone remains fixed, and only a lightweight linear head is trained to isolate the PAD information already present in the pretrained representation. 
%We report intra- and cross-dataset performance, along with accuracy–compute trade-offs, relative to two specialist PAD baselines. 
Results show that frozen foundation-model representations can support strong intra-dataset PAD performance with only a linear classifier, but this performance does not reliably transfer across datasets. Model scale is beneficial within several families, although the effect is not monotonic and is strongly mediated by architecture and pretraining. InternViT-6B achieves the lowest mean intra-dataset error, whereas CLIP ViT-B/32 offers the most favorable cross-dataset transfer–compute trade-off among the evaluated probes. 
These findings suggest that while pretrained representations contain PAD-relevant information, explicit adaptation remains necessary to address domain shift.
Code and evaluation protocols will be publicly released.
\href{https://www.idiap.ch/paper/foundationpad/}{https://www.idiap.ch/paper/foundationpad/}.
\end{abstract}

%% file: content/content.tex
%%%%%%%%% BODY TEXT
\section{Introduction}
Face recognition (FR) systems have improved substantially with deep learning~\cite{deng2019arcface,boutros2022elasticface}, yet remain vulnerable to presentation attacks (PAs)~\cite{OtroshiShahreza_IEEE-TIFS_2025}, such as printed photographs, replayed videos, and related spoofs presented to impersonate an enrolled identity and gain unauthorised access~\cite{zhang2020celeba,DBLP:journals/pr/FangDKK22,pasmino2023flickr} or avoid recognition (obfuscation).
Presentation attack detection (PAD) distinguishes bonafide presentations from attack samples before verification~\cite{DBLP:conf/cvpr/LiuJ018,DBLP:journals/pami/YuWQLLZ21,DBLP:journals/tbbis/YuLSXZ21,Fang_2022_WACV}.
Specialist deep neural network (DNN) approaches to PAD can achieve strong performance when trained and tested on the same dataset, but degrade under cross-dataset shifts in sensors, illumination, and attack types~\cite{DBLP:conf/cvpr/ShaoLLY19,DBLP:conf/aaai/ShaoLY20,DBLP:conf/cvpr/JiaZSC20,DBLP:journals/tbbis/WangWDG22,DBLP:journals/tcsv/YanZH22,fang2024face}.
The four MCIO PAD datasets (MSU-MFSD, CASIA-FASD, Replay-Attack, OULU-NPU) provide far fewer labelled samples than modern vision pretraining datasets such as ImageNet~\cite{deng2009imagenet}, limiting what end-to-end PAD models can learn from scratch~\cite{fang2023synthaspoof,fang2024face}.

\input{content/figures/linear_probe_arch}

One approach to addressing data scarcity and improving PAD detection is to leverage large-scale vision encoders and multimodal information.
Human learning is naturally multimodal, drawing on multiple sensory modalities to better understand and process new information.
Vision foundation models (VFMs) and multimodal large language models (MLLMs) pretrained on large-scale image or image--text data offer an alternative: generic representations that may already encode cues useful for spoof detection~\cite{radford2021learning,kirillov2023segment,oquab2023dinov2,chen2024internvl}.
% These approaches differ in how much of the backbone is updated and in training cost, making direct comparison difficult.

Recent works explore the use of foundation models (FMs) for PAD through multimodal architectures (e.g., FLIP~\cite{srivatsan2023flip}, I-FAS~\cite{zhang2025interpretable}, InstructFLIP~\cite{lin2025instructflip}), fine-tuning (e.g., with LoRA~\cite{hu2022lora} on CLIP~\cite{radford2021learning} in FoundPAD~\cite{ozgur2025foundpad}, and on DINOv2 with Registers~\cite{feng2026benchmarking}), and self-supervised pre-training without labels, i.e., FSFM~\cite{wang2025fsfm}.
By modifying the pre-trained weights of the VFM, these methods primarily assess adaptation performance rather than the general-purpose generalization properties of foundation models (FMs).

Recently, Gonzalez-Soler et al. investigated VFMs under zero-shot and linear-probing settings for PAD~\cite{gonzalez2025foundation}. 
In the linear-probing setting, they add only a trainable classification layer without altering the pretrained backbone. 
However, their experiments are limited to CLIP and DINO~\cite{oquab2023dinov2}.
Building on this idea, we systematically assess whether general-purpose foundation models (FMs) or other large-scale vision encoders encode sufficient PAD-relevant representations.

\textit{
The research question is: under a fixed linear-probing protocol on the PAD dataset, which general-purpose foundation model (FM) already separates bonafide from attacks intra-dataset, which of that performance survives cross-dataset transfer, and at what compute cost?
}

Keeping every backbone frozen isolates pretraining strategy and model scale from task-specific adaptation recipes such as LoRA or full fine-tuning.
We benchmark 24 pretrained vision encoders, including vision-only models and vision encoders from VLMs and MLLMs. % from InternViT~\cite{chen2024internvl}, CLIP~\cite{radford2021learning}, EVA-CLIP~\cite{sun2023eva}, Qwen~\cite{bai2023qwen}, DINOv2/v3~\cite{oquab2023dinov2,simeoni2025dinov3}, BEiT~\cite{bao2021beit}, ConvNeXt-V2~\cite{woo2023convnext}, SigLIP~\cite{zhai2023sigmoid}, ViT-MAE~\cite{he2022masked}, ViT-MSN~\cite{assran2022masked}, and under an identical protocol on the four MCIO datasets.
Two PAD specialist models (DeepPixBiS~\cite{george2019deep} and FSFM-FAS~\cite{wang2025fsfm}) serve as specialist baselines spanning classical convolutional neural network (CNN) PAD and recent ViT pretraining.
% We do not reproduce every published method, because our goal is representation benchmarking rather than chasing dataset-specific state-of-the-art.
% \red{Here, Just mention the main contributions; there is not enough context here to summarize the findings; this could move to discussions or conclusions after the experimental section}

\noindent Our main contributions are 
\begin{itemize}[nosep]
    \item a unified frozen-backbone linear-probing benchmark across a broad set of vision encoders (including vision encoders from VLMs and MLLMs) on standard PAD datasets.
    \item Besides intra-dataset evaluation, we also conduct cross-dataset evaluation and rank the models according to their performance in both settings.
    \item We also compare against representative specialist PAD baselines to contextualize when generic FMs justify their compute footprint.
\end{itemize} 

% Our main findings are threefold.
% First, backbone scale and pretraining objective strongly affect intra-dataset detection accuracy.
% InternViT-6B reaches sub-percent error on several dataset splits with only one trainable FC layer.
% Second, cross-dataset transfer remains difficult, while intra-dataset performance stays high for most FM, including large VFMs and specialist CNNs.
% Third, accuracy--compute trade-offs matter: compact specialists such as DeepPixBiS~\cite{george2019deep} remain competitive at a fraction of the inference cost of multi-billion-parameter backbones, while a mid-sized CLIP variant offers practical compromises.

% In summary, we (i)~introduce a unified frozen-backbone linear-probing benchmark across a broad FM zoo (including VLM and MLLMs) on standard PAD datasets.
% (ii)~report cross-dataset evaluation and efficiency analysis.
% (iii)~comparison against representative specialist PAD baselines to contextualize when generic FMs justify their compute footprint.

\section{Related Work}

% \red{May be merge both ?-- there is overlap in the different paragraphs; may be start with classifical pad and introduce cross database as the main challenge and so on ..
% }
\paragraph{Cross-dataset PAD and unseen-attack generalization.}
PAD methods exploit texture, colour, and frequency cues that distinguish live skin from printed or replayed media~\cite{DBLP:conf/cvpr/LiuJ018, DBLP:journals/tifs/LiLCWHK18, Fang_2022_WACV}.
Deep Neural Network (DNN) based models excel intra-dataset but degrade under cross-dataset shift~\cite{DBLP:journals/tbbis/WangWDG22,DBLP:journals/tcsv/YanZH22, fang2024face}.
Cross-dataset evaluation trains on one corpus and tests on another, exposing sensor and illumination shift~\cite{DBLP:conf/icmcs/LiuCDLZX22,fang2024face}.
Prior work addresses this through domain adaptation (DA), which aligns source and target feature spaces but requires target-domain data at training time~\cite{DBLP:journals/tifs/LiLCWHK18,DBLP:conf/icb/WangHSC19}, or domain generalization (DG), which trains on multiple labelled sources using adversarial or meta-learning objectives~\cite{DBLP:conf/cvpr/ShaoLLY19,DBLP:conf/aaai/ShaoLY20,DBLP:conf/cvpr/JiaZSC20,DBLP:conf/aaai/ChenYSDTLHJ21}.

Unseen-attack protocols hold out against presentation attacks (PAs) at test time. 
Leave-one-out (LOO; train on three MCIO corpora, test on the fourth) MICO~\cite{DBLP:conf/cvpr/ShaoLLY19} and limited source domain (LSD) protocols~\cite{feng2026benchmarking} (train on 2 and evaluate on the remainder) measure generalisation when multiple labelled sources are available at training time.
Auxiliary datasets, such as CelebA-Spoof~\cite{zhang2020celeba} and SynthA-Spoof~\cite{fang2023synthaspoof}, are injected during training to counterfeit data scarcity and to improve generalization capabilities~\cite{gonzalez2025foundation}. 

% Both families typically design task-specific architectures and training pipelines rather than asking what general-purpose pretrained vision encoders already provide.
% Our MCIO experiments use cross-dataset evaluation~\cite{bob2012,bob2017}, isolating the extent to which each frozen backbone transfers under minimal supervision.

% For example, benchmarks such as SiW-Mv2~\cite{xiaoguo2022MDFAS} make this explicit~\cite{gonzalez2025foundation}.

% Our study isolates cross-dataset transfer under linear probing and complements, rather than replaces, unknown-PA evaluation~\cite{fang2024face, gonzalez2025foundation}.

\paragraph{DNN PAD methods.}

Pixel-wise supervision~\cite{DBLP:conf/cvpr/LiuJ018} and depth-aware architectures~\cite{Fang_2022_WACV} train compact CNNs end-to-end on MCIO-scale data and can reach strong intra-dataset ACER. Instead of predicting a single live/spoof label or a dense depth map, the method predicts a pixel-wise binary map:
Bona fide image: every location in the supervision map is labeled as 1.
Attack image: every location is labeled as 0.
The network therefore learns local spoof artifacts through dense supervision while avoiding the complexity of depth estimation.
Frequency-domain and multi-scale designs further target print-and-replay artefacts~\cite{DBLP:journals/pr/FangDKK22}.
Patch-based formulations such as PatchNet~\cite{DBLP:conf/cvpr/WangLYL22} treat spoof detection as fine-grained local discrimination.
These networks remain useful compact deployment baselines because they are lightweight and dataset-tuned, but they often overfit acquisition conditions and degrade under cross-domain shift~\cite{DBLP:journals/tcsv/YanZH22,fang2024face}.

\paragraph{Foundation Models for PAD.}
Vision foundation models (VFMs) pretrained on web-scale image or image--text data offer an alternative to training PAD networks from scratch on small MCIO datasets~\cite{radford2021learning, oquab2023dinov2}.
FoundPAD~\cite{ozgur2025foundpad} adapts CLIP to PAD with Low-Rank Adaptation (LoRA)~\cite{hu2022lora}, updating a small subset of backbone weights while preserving most pretrained knowledge.
FSFM~\cite{wang2025fsfm} pretrains a vision transformer CLIP-B/16 backbone with face-spoofing-specific self-supervised learning (SSL) before downstream face anti-spoofing (FAS) fine-tuning.
Gonzalez-Soler~\textit{et al.}~\cite{gonzalez2025foundation} study zero-shot PAD with pretrained ViTs on the SiW-Mv2 dataset~\cite{xiaoguo2022MDFAS}, removing PAD-specific training entirely.

Later work extends this to ViTs through patch-level augmentation and patch-wise losses~\cite{watanabe2022pda}, a line that recent VFM studies~\cite{feng2026benchmarking} combine with large-scale self-supervised learning (SSL) backbones. 
SSL is a technique in which a model automatically generates its own training labels from the raw data, rather than relying on manual human annotations.

% Feng~\textit{et al.}~\cite{feng2026benchmarking} systematically compare vision-only foundation models under severe cross-dataset FAS protocols, including MICO leave-one-out (LOO; train on three MCIO corpora, test on the fourth) and limited source domains (LSD, train on two, test on the remainder).
Feng~\textit{et al.}~\cite{feng2026benchmarking} compare frozen backbones and find that self-supervised ViTs outperform supervised CNNs and ImageNet ViTs, with DINOv2 augmented with register tokens~\cite{darcet2024registers} yielding the most consistent cross-domain accuracy.
They further show that PAD-specific adaptation, including Face Anti-Spoofing Data Augmentation (FAS-Aug)~\cite{cai2024fasaug}, patch-wise data augmentation (PDA)~\cite{watanabe2022pda}, and attention-weighted patch loss (APL)~\cite{watanabe2022pda}, on a fine-tuned DINOv2 backbone closes much of the gap to large vision--language FAS systems at a fraction of the parameter count.
Other recent cross-dataset methods explore prompt learning~\cite{liu2024cfpl}, frequency-domain modelling~\cite{cao2025towards}, domain--modality alignment~\cite{yang2025dadm}, and optimal-transport adaptation~\cite{li2025optimal}.
Together, these approaches map a spectrum from zero-shot inference~\cite{gonzalez2025foundation}, through lightweight probing, task-adaptive LoRA~\cite{ozgur2025foundpad}, to heavily engineered cross-domain training~\cite{liu2024cfpl, cao2025towards, li2025optimal, wang2025fsfm, yang2025dadm, feng2026benchmarking}.

% Our work differs in scope and protocol.
% FoundPAD~\cite{ozgur2025foundpad} tests task-adaptive LoRA on CLIP.
% We freeze every backbone to measure general-purpose representations before any encoder update.
% Feng~\textit{et al.}~\cite{feng2026benchmarking} benchmark VFMs under MICO/LSD with backbone fine-tuning, PAD-specific augmentations, and patch losses, reporting AUC and HTER.
% We use Bob single-source cross-dataset evaluation, dev equal error rate (EER) thresholding with ACER, a broader encoder zoo (including multimodal towers and billion-scale models), explicit cross-dataset aggregates, and accuracy--compute analysis against specialist baselines.
% FSFM~\cite{wang2025fsfm} is included as one end-to-end specialist baseline trained under the same splits and ACER metric as our frozen encoders.
% We do not reproduce every published pipeline.
% Instead, we benchmark frozen VFMs under one protocol and compare against two end-to-end specialist baselines (DeepPixBiS, FSFM-FAS) that span classical CNN PAD and recent face-spoofing pretraining.

\section{Approach}

\subsection{Presentation Attack Detection Baseline}

To contextualize the frozen-probing results, we compare them with two end-to-end specialized PAD models trained or fine-tuned directly for the task. 
These specialist baselines span classical CNN-based supervision and SSL-pretrained ViT fine-tuning.
DeepPixBiS~\cite{george2019deep} is a strong DNN approach trained with pixel-wise supervision.

FSFM-FAS~\cite{wang2025fsfm} is a ViT-B/16 fine-tuned after SSL pretraining on the VGGFace2 dataset~\cite{cao2018vggface2}.
Wang et al. report that FSFM-FAS outperforms the few-shot FAS approach of Huang et al.~\cite{huang2022adaptive} on their benchmarks~\cite{wang2025fsfm}.

% Both are fully fine-tuned on each MCIO corpus under the same Bob protocol splits (train, dev, test) and ACER metric as our frozen encoders.
% InternVL-14B zero-shot scores provide a zero-shot comparator, indicating how a general-purpose vision--language model (VLM) performs without any PAD-specific readout or training.
% These specialist baselines span classical CNN and face-spoofing SSL ViT training, against which to interpret frozen VFM probing.

\subsection{Frozen FM Backbone with Linear Probing}

We evaluate frozen vision encoders with linear probing (e.g., for ViT see Fig.~\ref{fig:linear-probe-arch}).
Each cropped face image~$x$ is passed through a pretrained vision backbone whose weights remain fixed throughout training and testing.
The frozen backbone outputs a global image embedding $h \in \mathbb{R}^{d}$,
whose extraction rule depends on the encoder family
(CLS token, post-layer-norm CLS, global average pooling, projected CLIP image embedding, or mean-pooled patch tokens);
see the Feature column in~\cref{tab:indomain-acer}.
Only a single fully connected layer maps~$h$ to a scalar logit~$z = w^{\top} h + b$.%, with~$d{+}1$ trainable parameters in total.
Training minimizes binary cross-entropy on~$z$ with bonafide labels $y{=}1$ and attack labels~$y{=}0$.
At inference, the bonafide probability is computed as: $p=\sigma(z)$.

% Freezing every backbone serves three purposes.
% First, it isolates pretraining strategy and model scale from task-specific adaptation recipes, such as LoRA~\cite{ozgur2025foundpad} or full fine-tuning, thereby providing a reproducible lower bound on what each encoder affords before any weight updates.
% Second, this approach equalizes trainable capacity across all 22 frozen models. After each frozen backbone, there is a fully connected trainable layer.  
% Consequently, differences in ACER reflect representation quality and FC capacity.
% Third, it enables fair comparison against published task-adaptive methods.
% Our frozen scores indicate how much PAD signal is already present in general-purpose encoders, while methods such as FoundPAD~\cite{ozgur2025foundpad} measure the incremental gain from backbone updates.

% The end-to-end pipeline is: MTCNN face crop ($224{\times}224$) $\rightarrow$ HuggingFace image processor $\rightarrow$ frozen encoder $\rightarrow$ linear probe $\rightarrow$ dev-split threshold (EER) $\rightarrow$ test ACER (Sec.~\ref{sec:experiments}).

\subsection{Vision Encoders}

\input{content/tables/encoder_taxonomy}

\Cref{tab:encoder-taxonomy} lists the 24 frozen backbones grouped by family and scale. 
ViT-style encoders (CLIP, DINOv2/v3, BEiT, SigLIP, ViT-MAE/MSN, InternViT) use the final CLS token; 
Qwen2 uses mean pooling over patch tokens; ConvNeXt-V2 uses global average pooling.

\textbf{Multimodal LLM Vision Encoders.} 
Multimodal large language models (MLLMs) integrate vision and language within a unified architecture, formally $\mathcal{M}_{\mathrm{MLLM}} = \mathcal{V} \oplus \mathcal{L} \oplus \mathcal{O}$, where $\mathcal{V}$ denotes the vision encoder, $\mathcal{L}$ the language model, and $\mathcal{O}$ optional other modalities.
InternVL~\cite{chen2024internvl} follows a ViT-MLP-LLM paradigm with InternViT (300M, 6B parameters) as the vision tower and Qwen3~\cite{yang2025qwen3} as the language model; a multi-layer projector bridges visual representations to the language embedding space. 
The full pipeline is optimized end-to-end on large image-text datasets through contrastive alignment (matching image-caption pairs) and language-model loss (generating coherent text from projected visual tokens).
QwenVL's vision encoder~\cite{bai2023qwen} spans 0.5B--72B parameters and employs a mixture-of-experts design that activates a subset of experts per token for efficient computation.

\textbf{Vision-language Contrastive Models.} 
CLIP~\cite{radford2021learning} (B/32, L/14, H/14) learns aligned image-text representations through contrastive learning on 400 million web-crawled image-caption pairs, enabling zero-shot transfer to tasks for which it was not explicitly trained. 
The model variants differ in scale (Base, Large, Huge) and patch size (32 vs. 14 pixels), where larger patches reduce computational cost while potentially discarding fine-grained details.
SigLIP~\cite{zhai2023sigmoid} (Base, Large) replaces the softmax in CLIP with a sigmoid loss, allowing independent processing of each image-text pair and more efficient batch-size scaling, which is particularly beneficial for large-scale training.
EVA-CLIP-8B~\cite{sun2023eva} pushes CLIP to larger scale by initializing the vision encoder with masked-image pretraining (similar to BERT for text) before contrastive fine-tuning on web-scale data.

\textbf{Self-supervised Vision Encoders.} 
BEiT~\cite{bao2021beit} (Base, Large) performs masked image modeling by predicting discrete visual tokens obtained from a pre-trained discrete VAE for masked patches, excelling at semantic understanding tasks.
DINOv2~\cite{oquab2023dinov2} combines self-distillation (learning without labels) with iBOT~\cite{zhou2021ibot} (masked modeling in latent space) and is trained on a curated 142M-image dataset (LVD-142M), built by augmenting ImageNet-22k~\cite{deng2009imagenet} and Google Landmarks~\cite{weyand2020google} with visually similar web images.
DINOv3~\cite{simeoni2025dinov3} addresses scaling limitations of its predecessor with a modified training recipe (constant hyperparameters replacing cosine schedules) and Axial RoPE for positional embeddings, resulting in improved spatial precision and robustness; it is trained on the larger, more diverse LVD-1689M dataset.
MSN~\cite{assran2022masked} learns by matching heavily masked views to prototypes (learned visual centers) derived from unmasked views, though it performs near chance in our setting, suggesting that not every self-supervised objective yields PAD-relevant features.

\textbf{Generative masked image modeling.} 
ViT-MAE~\cite{he2022masked} (Base, Huge) directly reconstructs RGB pixel values of masked patches without relying on a discrete tokenizer, learning rich representations through a simple and scalable reconstruction objective.

\textbf{Modern CNNs.} 
ConvNeXt-V2~\cite{woo2023convnext} (Base, Huge) modernizes the classic ResNet architecture with transformer-inspired design elements such as large-kernel depthwise convolutions, inverted bottlenecks, layer-scale, and GELU activations while preserving the strong locality bias of convolutions that CNNs naturally possess.

\section{Experiments}
\label{sec:experiments}

\subsection{Datasets and Protocol}

We evaluate only the single-source cross-dataset protocol, where a model is trained on one source dataset and evaluated on a different target dataset. Multi-source protocols such as leave-one-out (LOO) and limited-source-domain (LSD) evaluation are not considered.
We evaluate on four widely used MCIO datasets~\cite{wen2015face, costa2016replay, patel2016secure, boulkenafet2017OULU} for intra-dataset ($M \rightarrow M$, $C \rightarrow C$, $I \rightarrow I$, $O \rightarrow O$) and cross-dataset ($M \rightarrow C$, $M \rightarrow I$, $M \rightarrow O$, ..., $O \rightarrow$ ...).
MSU-MFSD\footnote{M: Protocol \texttt{protocol} (all folds)
\href{https://files.pythonhosted.org/packages/45/18/3e2b324468d3cdc290ac08b73f8972fa2aa87bc1071605057acfe3dd9900/bob.db.msu_mfsd_mod-2.2.9.zip}{shorturl.co/msu-mfsd-mod-2.2.9}}
(\textbf{M}) includes videos of printed attacks, as well as replay attacks generated with a mobile phone and a tablet.
%\url{https://files.pythonhosted.org/packages/45/18/3e2b324468d3cdc290ac08b73f8972fa2aa87bc1071605057acfe3dd9900/bob.db.msu_mfsd_mod-2.2.9.zip}.} 
CASIA-FASD\footnote{C: Protocol \texttt{grandtest}  
\href{https://www.idiap.ch/software/bob/data/bob/bob.pad.face/pad-face-casia-fasd-0b07ea45.tar.gz}{shorturl.co/casia-fasd-0b07ea45}}
%\url{https://www.idiap.ch/software/bob/data/bob/bob.pad.face/pad-face-casia-fasd-0b07ea45.tar.gz}.}
(\textbf{C}) includes warped-photo, cut-photo, and replay attacks under varying illumination.
Replay-Attack\footnote{I: Protocol \texttt{grandtest} 
\href{https://www.idiap.ch/software/bob/data/bob/bob.pad.face/pad-face-replay-attack-aca6b46f.tar.gz}{shorturl.co/replay-attack-aca6b46f}}
%\url{https://www.idiap.ch/software/bob/data/bob/bob.pad.face/pad-face-replay-attack-aca6b46f.tar.gz}.} 
(\textbf{I}) provides high-resolution print-and-replay spoofs on a fixed display setup.
OULU-NPU\footnote{O: Protocol \texttt{Protocol\_1} 
\href{https://www.idiap.ch/software/bob/data/bob/bob.pad.face/pad-face-oulunpu-7bfb90c9.tar.gz}{shorturl.co/oulunpu-7bfb90c9}}
%\url{https://www.idiap.ch/software/bob/data/bob/bob.pad.face/pad-face-oulunpu-7bfb90c9.tar.gz}.} 
(\textbf{O}) comprises six mobile phone sensors and multiple print-and-replay attack types, offering and the strongest sensor diversity.

Together, M/C/I/O cover laboratory and mobile capture, multiple PA, and the cross-dataset shift typical of PAD benchmarking~\cite{DBLP:conf/icmcs/LiuCDLZX22,fang2024face}.
Per-split (train, dev, test) frame counts are in~\cref{tab:mcio-ba} and Appendix~\cref{app:dataset-counts}.

Intra-dataset results are reported in~\cref{tab:indomain-acer,tab:cross-dataset-representative,tab:cross-dataset-summary} of this evaluation.
cross-dataset results are reported in~\cref{tab:cross-dataset-representative,tab:cross-dataset-summary}.%, where details for the calculation can be found in~\cref{eq:mean-acer-all} and~\cref{eq:mean-acer-offdiag}.
% All experiments follow a cross-dataset setting: Models are trained on one MCIO corpus and evaluated on the same or a different corpus.
% For each encoder, we perform cross-dataset evaluation across all four corpora: the model trains on corpus~$s$ only and is tested on corpus~$t$, yielding twelve cross-corpus transfer pairs plus four intra-dataset pairs (\cref{eq:mean-acer-all}, \cref{eq:mean-acer-offdiag}).
% Triple- and double-source MCIO protocols are widely used in the literature~\cite{DBLP:conf/aaai/ChenYSDTLHJ21,DBLP:conf/icmcs/LiuCDLZX22,fang2024face}.
% We focus on single-source training to isolate how much each pretrained backbone transfers when it is fit with minimal supervision on a single corpus at a time.
% \begin{table}[htpb]
% \centering
% \caption{MCIO datasets frame counts per split.}
% \label{tab:mcio-splits}
% \setlength{\tabcolsep}{4pt}
% \begin{tabular}{lrrr}
% \toprule
% Dataset & Train & Dev & Test \\
% \midrule
% MSU-MFSD (M)      &  3{,}196 &  3{,}037 &  2{,}398 \\
% CASIA-FASD (C)    &  3{,}600 &  1{,}200 &  7{,}200 \\
% Replay-attack (I) &  7{,}199 &  7{,}200 &  9{,}600 \\
% Oulu-NPU (O)      & 24{,}000 & 17{,}999 & 12{,}000 \\
% \midrule
% Total             & 37{,}995 & 29{,}436 & 31{,}198 \\
% \bottomrule
% \end{tabular}
% \end{table}
\begin{table}[htbp]
\centering
\caption{MCIO frame counts (B/A = bonafide/attack).}
\label{tab:mcio-ba}
\setlength{\tabcolsep}{3pt}
\footnotesize
\begin{tabular}{llll}
\toprule
 & Train & Dev & Test \\
\midrule
M & 3196 (798/2398)   & 3037 (759/2278)   & 2398 (599/1799) \\
C & 3600 (900/2700)   & 1200 (300/900)    & 7200 (1800/5400) \\
I & 7199 (1200/5999)  & 7200 (1200/6000)  & 9600 (1600/8000) \\
O & 24000 (4800/19200)& 17999 (3600/14399)& 12000 (2400/9600) \\
\bottomrule
\end{tabular}
\end{table}

\textbf{Pre-processing.}
We use MTCNN\footnote{MTCNN weights \href{https://github.com/timesler/facenet-pytorch}{github.com/timesler/facenet-pytorch}.}~\cite{MTCNN} face detection preprocessing pipeline to crop faces in a size of $224{\times}224$ pixels from the video.
Up to 20 frames are sampled evenly spaced across the video.
Fig.~\ref{fig:mcio-samples} shows representative bonafide and attack crops from each MCIO corpus after this step.
Each FM is paired with its image processor for resizing, normalization, and family-specific preprocessing.
The most common input size is $224{\times}224$, although it varies across models: InternViT uses $448{\times}448$ inputs (Appendix~\ref{app:internvit}), while SigLIP-Large uses $256{\times}256$ inputs.

\input{content/figures/mcio_samples}

\textbf{Models.}
\Cref{tab:encoder-taxonomy} and~\cref{tab:indomain-acer} list 24 frozen vision backbones.
Only the linear probe is trained.
Two end-to-end specialist PAD models (DeepPixBiS~\cite{george2019deep} and FSFM-FAS~\cite{wang2025fsfm}) serve as specialist baselines under the same datasets and ACER metric (fully fine-tuned, not linearly probed).

\textbf{Implementation details.}
Training minimizes \texttt{BCEWithLogitsLoss} on binary logits with class-balanced mini-batches (\texttt{WeightedRandomSampler}).
We optimize the probe with Adam ($\mathrm{lr}{=}10^{-4}$, weight decay $10^{-6}$), batch size~64, and a cosine learning-rate schedule (minimum $10^{-6}$).
Each run trains for up to 5--20 epochs with early stopping on validation loss (patience~8, minimum 5 epochs).
Training-time augmentation applies random horizontal flip and color jitter (brightness, contrast, and saturation $0.15$).
% Specialist PAD baselines (DeepPixBiS, FSFM\_FAS) are trained end-to-end under their respective published configurations with all weights updated.
We use PyTorch Lightning as the main training framework. 
The Bob toolkit \cite{bob2012, bob2017} provides the software library to calculate the EER and ACER metrics.

\subsection{Evaluation metrics}

We report the \emph{Average Classification Error Rate} (ACER) following ISO/IEC~30107-3~\cite{ISO301073} and standard PAD practice~\cite{DBLP:conf/icmcs/LiuCDLZX22,Fang_2022_WACV,fang2024face}.
ACER is the mean of attack and bonafide error rates at a threshold $\tau$ tuned on the development split and applied unchanged to the test split:
\begin{equation}
  \mathrm{ACER}(\tau)
  = \tfrac{1}{2}\bigl(\mathrm{APCER}(\tau) + \mathrm{BPCER}(\tau)\bigr).
  \label{eq:acer}
\end{equation}
Full APCER/BPCER definitions are in Appendix~\ref{app:metrics}.
All reported ACER values are in percent (\%) (lower is better).
The operating threshold $\tau$ is selected on the development split by minimizing equal error rate (EER criterion), then applied to the test split.
This dev-tuned thresholding is standard in PAD evaluation because it balances attack and bonafide error rates before reporting cross-corpus transfer.

Let $\mathcal{C}=\{\mathrm{M},\mathrm{C},\mathrm{I},\mathrm{O}\}$ denote the datasets and $A_{s,t}$ the test ACER (Eq.~\eqref{eq:acer}) when the model is trained on corpus~$s$ and evaluated on corpus~$t$.
The full cross-dataset evaluation results appear in Appendix~\ref{app:cross-dataset-generalizability} (Tables~\ref{tab:cross-dataset-acer}--\ref{tab:cross-dataset-acer-cont3}).
Table~\ref{tab:cross-dataset-representative} shows two representative models in the main paper.
We report two unweighted cross-dataset aggregates:
\begin{align}
  {\mathrm{ACER}}_{\mathrm{all}}
  &= \frac{1}{|\mathcal{C}|^2}
     \sum_{s \in \mathcal{C}} \sum_{t \in \mathcal{C}} A_{s,t},
  \label{eq:mean-acer-all} \\
  {\mathrm{ACER}}_{\mathrm{\colorbox{green!30}{cross}}}
  &= \frac{1}{|\mathcal{C}|(|\mathcal{C}|-1)}
     \sum_{s \in \mathcal{C}} \sum_{\substack{t \in \mathcal{C}\\ t \neq s}} A_{s,t}.
  \label{eq:mean-acer-offdiag}
\end{align}
\Cref{eq:mean-acer-all} averages all sixteen train/test pairs (including the intra-dataset diagonal).
\Cref{eq:mean-acer-offdiag} averages the twelve off-diagonal transfer pairs only.
\Cref{tab:cross-dataset-summary} reports Diag.\ (mean diagonal), Cross (mean off-diagonal), and Gap $=$ Cross $-$ Diag for each model.
% A large Gap indicates strong intra-dataset performance that does not survive cross-corpus transfer (e.g., DeepPixBiS: 3.4\% Diag., 43.1\% Cross, Gap 39.7).

% \section{Results}
% \label{sec:results}

\subsection{Intra-dataset (ID) and Cross-dataset (CD) Performance}

\textbf{Intra-Dataset (ID).} Table~\ref{tab:indomain-acer} reports test ACER in percent for intra-dataset training on MCIO datasets.
The upper block lists frozen FMs in different sizes probed linearly.
The lower block lists specialist baselines and a zero-shot MLLM InternVL-14B (containing InternViT-300M) models.
A common trend is that larger foundation models (FMs) generally outperform smaller ones, which is also observed by Zhai et al.~\cite{zhai2022scaling}. 
% InternViT-6B reaches the lowest average intra-dataset ACER (1.6\%), followed by the FM DINOv3 (5.7\%), ahead of the baselines DeepPixBiS (3.4\%) and FSFM-FAS (4.2\%).
InternViT-6B achieves the lowest average intra-dataset ACER (1.6\%). 
Among frozen foundation models, DINOv3-giant achieves the second-lowest average error (5.7\%). However, the specialist baselines DeepPixBiS (3.4\%) and FSFM-FAS (4.2\%) still outperform most frozen probes.
DeepPixBiS does so with only ${\sim}$3M trainable weights, while InternViT has the largest linear probe with ${\sim}$3k trainable weights.
The InternVL-14B zero-shot model averages 28.8\% ACER, far above even weak linear probes, indicating that generic VLM inference without a task-aligned readout is insufficient.

% Specialist models show corpus-specific peaks and troughs (e.g., DeepPixBiS on Replay-Attack: 0.0\%) but trade smaller frozen encoders' peak performance for higher variance.
% InternViT uses $448{\times}448$ inputs whereas most other encoders use $224{\times}224$.
% We report both under each model's native processor rather than retraining all backbones at one resolution.

\input{content/tables/indomain_acer}

% \subsection{Cross-domain (CD) performance}
% \label{sec:cross-dataset}

\textbf{Cross-Dataset (CD).} Full MCIO cross-dataset test ACER results for all benchmarked models are in Appendix~\ref{app:cross-dataset-generalizability}.
Table~\ref{tab:cross-dataset-representative} shows InternViT-6B and DeepPixBiS as representative cases, but CD (off-diagonal) entries remain high for nearly every model.
When InternViT-6B is trained on MSU-MFSD, it reaches 4.4\% on the diagonal but 14.7--45.3\% on the three held-out corpora.
DeepPixBiS trained on the same source shows a similar gap: 7.9\% intra-dataset versus 36.5--49.7\% off-diagonal.
Even strong ID (diagonal cells) does not transfer reliably: DeepPixBiS trained on Replay-Attack reaches 0.0\% intra-dataset yet 47.4--61.3\% on other test corpora.
InternViT-6B improves many pairs relative to InternViT-300M (${\mathrm{ACER}}_{\mathrm{all}}$: 41.8\% $\rightarrow$ 24.4\%, ${\mathrm{ACER}}_{\mathrm{cross}}$: 47.7\% $\rightarrow$ 32.0\%) and follows the pattern that the larger model has better frozen features.
% Cross-dataset performance remains uneven.
% The specialist CNN baseline follows the same pattern, so low intra-dataset ACER is not a reliable proxy for cross-dataset PAD.

\input{content/tables/cross_dataset_representative}

% Transfer is asymmetric across train:test pairs.
% Training on Replay-Attack (I) yields the lowest intra-dataset ACER for several specialists (DeepPixBiS: 0.0\%; InternViT-6B: 0.06\%) but the highest off-diagonal error when tested on CASIA-FASD (DeepPixBiS I$\rightarrow$C: 61.3\%; InternViT-6B I$\rightarrow$O: 50.3\%).
% CASIA-FASD remains difficult as both train and test: cross-dataset evaluation above 30\% for most models.
% % OULU-NPU benefits from the largest training set (${\sim}$24\,k frames) but its six-sensor diversity creates substantial shift when models trained on laboratory corpora are tested on mobile captures, and vice versa.

\input{content/tables/cross_dataset_summary}

% \subsection{Scaling within encoder families}
% \label{sec:family-scaling}

% Table~\ref{tab:cross-dataset-summary} summarizes mean cross-dataset ACER (${\mathrm{ACER}}_{\mathrm{all}}$) by model size within each encoder family.
% Scaling trends are family-dependent.
% DINOv2 base$\rightarrow$giant reduces ${\mathrm{ACER}}_{\mathrm{all}}$ from 35.5\% to 31.9\%.
% DINOv3 base$\rightarrow$giant reduces it from 33.2\% to 32.7\%.
% CLIP B/32$\rightarrow$L/14$\rightarrow$H/14 moves from 26.8\% to 28.2\% to 31.6\%.
% SigLIP base$\rightarrow$large moves from 39.5\% to 36.3\%.
% InternViT 300M$\rightarrow$6B moves from 41.8\% to 24.4\%.
% Larger variants generally improve cross-dataset ACER, but gains diminish and can reverse within a family (CLIP H/14 underperforms L/14 on ${\mathrm{ACER}}_{\mathrm{all}}$).
% InternViT shows the steepest scaling benefit, consistent with its multimodal pretraining and higher input resolution.

% \begin{figure*}[t]
%   \centering
%   \includegraphics[width=\linewidth]{pad/efficiency/fig_family_scaling}
%   \caption{Mean cross-dataset ACER (${\mathrm{ACER}}_{\mathrm{all}}$, Eq.~\eqref{eq:mean-acer-all}) vs.\ model size within encoder families (DINOv2, DINOv3, CLIP, SigLIP, InternViT). Lower ACER is better.}
%   \label{fig:family-scaling}
% \end{figure*}

% \subsection{Generalization gap analysis}
% \label{sec:gen-gap}

%\Cref{tab:cross-dataset-summary} %reports Diag., Cross, and Gap $=$ Cross $-$ Diag for every benchmarked model.
\textbf{Gap.} In \Cref{tab:cross-dataset-summary}, most models show moderate-to-high CD error despite varying ID performance.
InternViT-6B achieves the lowest intra-dataset ACER (1.6\% Diag.) but a 30.3\% ACER gap, illustrating that strong linear separability within a corpus does not imply CD robustness.
ViT-MSN occupies the opposite corner: near-chance intra-dataset ACER (${>}46\%$) with a small Gap (${\sim}$9--10\%), indicating uniformly poor representations rather than overfitting to a single corpus.
DeepPixBiS and FSFM-FAS combine low ID error with large gaps (${>}33$ points), indicating that fine-tuning can sharpen ID decision boundaries without improving CD generalization.
\subsection{Accuracy vs compute trade-offs}
\label{sec:efficiency}
\begin{figure*}[t]
  \centering
  \includegraphics[width=\linewidth]{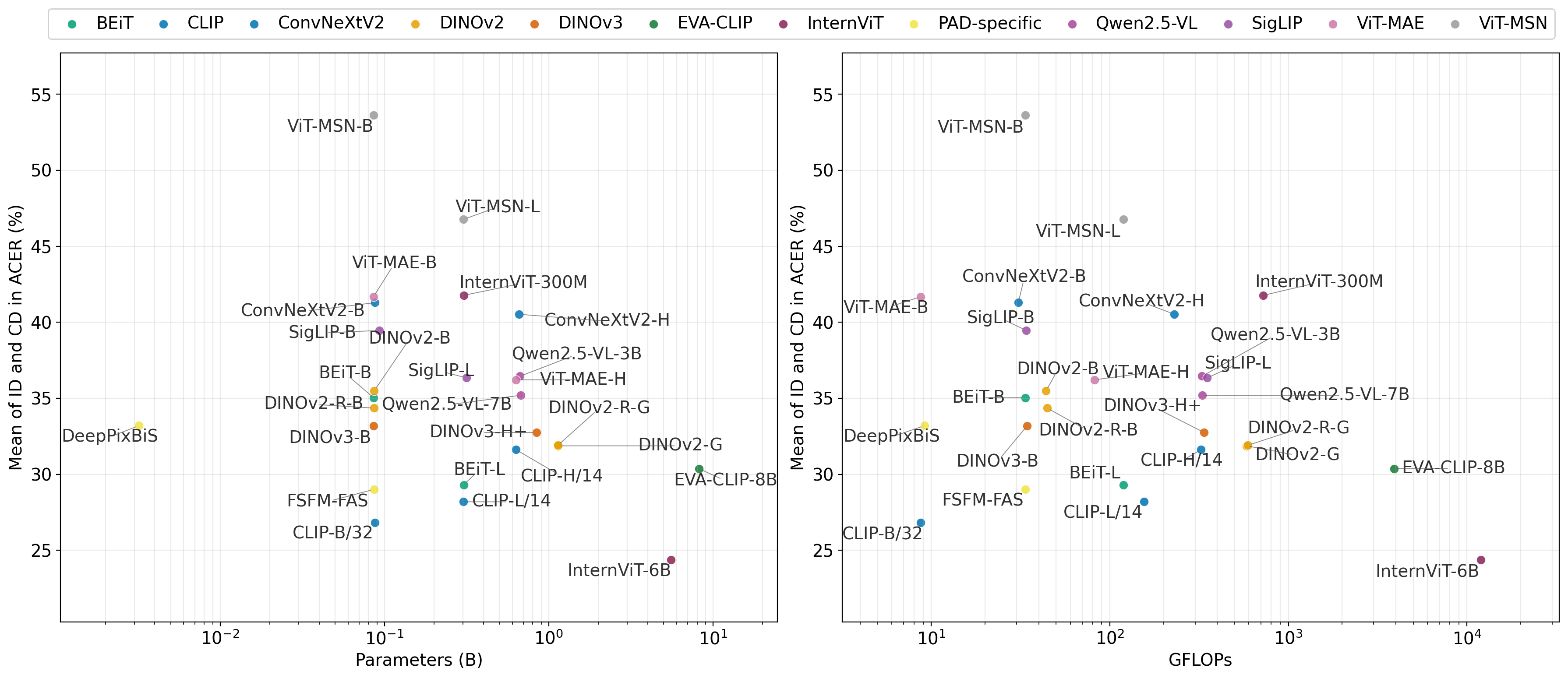}
  \caption{
  Comparison of the overall classification performance (intra-dataset and cross-dataset transfer measured  in ${\mathrm{ACER}}_{\mathrm{all}}$ (\%), 
  \Cref{eq:mean-acer-all})  vs.\ GFLOPs per model.
    InternViT-6B has the second largest number of parameters and highest GFLOPs, while also achieving the best overall detection performance. Note that InternViT-6B processes images at approximately twice the resolution (see \cref{tab:encoder-taxonomy}), which is typically associated with a higher number of GFLOPs~\cite{radford2021learning}.
    Although CLIP-B/32 has fewer GFLOPs than DeepPixBiS, its model size is much larger.
    CLIP-B/32 is also outperforming FSFM-FAS (backbone: CLIP-B/16), the specialized foundation model, on detection accuracy.
  {
  \scriptsize 
  Points are colored by model family.
  Lower ACER is better.
  }
  }
  \label{fig:efficiency-compute}
\end{figure*}

Fig.~\ref{fig:efficiency-compute} summarizes the accuracy--compute trade-off on the full MCIO cross-dataset evaluation.
Larger backbones and higher-GFLOP encoders tend toward lower ${\mathrm{ACER}}_{\mathrm{all}}$.
No single encoder dominates every budget.
Mid-sized DINOv3 and CLIP variants offer practical compromises.
CLIP ViT-B/32 is a notable outlier on this plot: with only ${\sim}$90\,M frozen parameters, it ranks second on ${\mathrm{ACER}}_{\mathrm{all}}$ (26.8\%) and attains the lowest mean off-diagonal ACER in~\cref{tab:cross-dataset-summary} (31.6\%), ahead of much larger CLIP and DINO variants.
This pattern is plausible because contrastive image--text pretraining on diverse web-scale data encourages visual features that are less tied to a single acquisition pipeline~\cite{radford2021learning}, and frozen CLIP backbones have previously ranked among the strongest cross-dataset extractors in VFM-FAS benchmarks~\cite{srivatsan2023flip,liu2024cfpl,feng2026benchmarking}.
Within the CLIP family, B/32 also outperforms ViT-L/14 and ViT-H/14 on cross-corpus transfer (31.6\% vs.\ 34.0\% and 38.7\% mean off-diagonal ACER), suggesting that coarser $32{\times}32$ patch tokenisation can trade some intra-dataset separability for robustness to MCIO shift under a linear readout.
The right panel of~\cref{fig:efficiency-compute} also highlights that CLIP ViT-B/32 and the compact CNN specialist DeepPixBiS~\cite{george2019deep} require almost the same forward-pass cost (${\sim}$8.7 vs.\ ${\sim}$9.2 GFLOPs at $224{\times}224$), yet frozen CLIP probing yields markedly lower cross-dataset error (${\mathrm{ACER}}_{\mathrm{all}}$ 26.8\% vs.\ 33.2\%; mean off-diagonal 31.6\% vs.\ 43.1\%).
Similar inference budgets do not guarantee similar transfer: a task-trained specialist can match intra-dataset performance, but a frozen web-scale encoder offers stronger cross-corpus robustness at comparable compute.

\subsection{Embedding Structure}
\label{sec:embeddings}
\begin{figure}[tp]
  \centering
  \begin{subfigure}[t]{\linewidth}
    \centering
    \includegraphics[width=\linewidth]{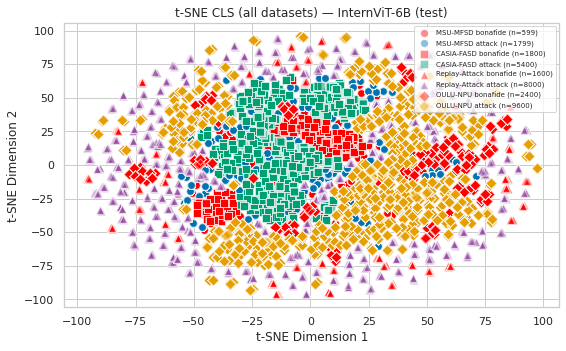}
    \caption{InternViT-6B}
    \label{fig:tsne-internvit-6b}
  \end{subfigure}
  \vspace{0.4em}
  \begin{subfigure}[t]{\linewidth}
    \centering
    \includegraphics[width=\linewidth]{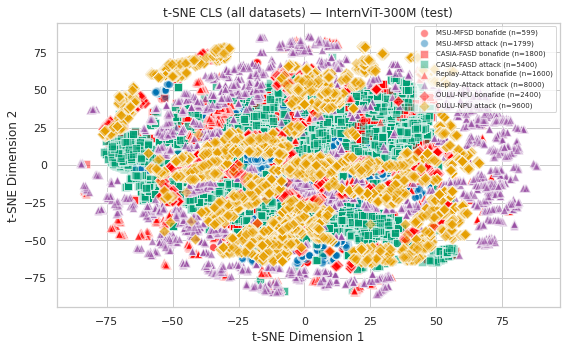}
    \caption{InternViT-300M}
    \label{fig:tsne-internvit-300m}
  \end{subfigure}
  \caption{t-distributed stochastic neighbor embedding (t-SNE) of frozen embeddings (MCIO test, all corpora pooled). Bonafide: red. Attacks share the corpus hue.  Umap plots in Appendix~\cref{app:tsne-cls}.}
  \label{fig:tsne-main}
\end{figure}
Fig.~\ref{fig:tsne-main} visualizes embedding structure for the two InternViT FMs.
InternViT-6B forms more compact clusters than InternViT-300M, consistent with lower intra-dataset ACER. Yet, both retain strong dataset-specific structure that aligns with the cross-dataset error in~\cref{tab:cross-dataset-summary}.
% For the remaining models, the corresponding visualizations are provided in t-SNE~\cref{app:tsne-cls} and UMAP~\cref{app:umap-cls}.

\subsection{Discussion}

Our results reveal a clear disconnect between intra-dataset performance and cross-dataset robustness. 
Models achieving near-perfect intra-dataset performance often exhibit substantial degradation under domain shift. 
InternViT-6B achieves the lowest error in intra-dataset settings, yet its mean cross-dataset ACER remains high (\cref{tab:cross-dataset-summary}), underscoring the difficulty of domain generalization even with massive backbones.
FSFM-FAS, despite being pretrained specifically for face anti-spoofing, does not yield better PAD features than generic large-scale pretraining: it averages 4.2\% intra-dataset ACER, higher than both InternViT-6B and DeepPixBiS.
DeepPixBiS remains corpus-specific, strong on MSU-MFSD and Replay-Attack, weak on CASIA-FASD and OULU-NPU, while mid-sized frozen encoders such as DINOv3-giant and CLIP ViT-L/14 trade peak performance for lower variance across datasets (Table~\ref{tab:indomain-acer}).

Pre-training objective appears at least as important as parameter count. 
Self-supervised and vision-language encoders (DINOv2/v3, BEiT, SigLIP, CLIP, EVA-CLIP) consistently outperform similarly sized ViT-MAE, ViT-MSN, and ConvNeXt-V2 variants on both intra- and cross-dataset ACER (Table~\ref{tab:cross-dataset-summary}). 
ViT-MSN performs near chance ($\mathrm{ACER}_{\mathrm{all}}{>}53\%$) despite its ViT architecture, suggesting that not every large-scale pretraining objective yields PAD-relevant separable features. 
This aligns with the view that PAD benefits from encoders that preserve stable, localisable spatial structure. 
Feng~\textit{et al.}~\cite{feng2026benchmarking} argue that standard ViTs can misallocate attention to background regions, whereas DINOv2 with register tokens produces smoother feature maps better suited to fine-grained artefact detection~\cite{darcet2024registers}. 
Under our frozen linear-probe setting, the DINO and BEiT families already outperform reconstruction-based (ViT-MAE) and prototype-based (ViT-MSN) objectives, suggesting that self-distillation and discrete-token prediction preserve more PAD-relevant local structure.

Among families with multiple sizes (DINOv2, DINOv3, CLIP, SigLIP, InternViT), larger variants generally reduce $\mathrm{ACER}_{\mathrm{all}}$, though gains are family-dependent and non-monotonic. 
InternViT's superior performance may be partly attributed to its $448{\times}448$ pretraining resolution (Appendix~\ref{app:internvit}), yet this does not eliminate the cross-dataset gap (Table~\ref{tab:cross-dataset-summary}).

These frozen-probe scores represent a lower bound on what each backbone can achieve. 
Feng~\textit{et al.}~\cite{feng2026benchmarking} demonstrate that fine-tuning SSL backbones with spoof-specific augmentation (FAS-Aug) and patch supervision (PDA/APL) substantially reduces cross-dataset error on the same MCIO corpora. 
The large generalization gaps we report therefore reflect both limited source diversity in single-corpus training and the absence of task-specific adaptation, rather than an inherent limitation of pretrained vision models.

\section{Conclusions}
We benchmarked vision encoders of 24 frozen MLLMs, VFMs, and specialized networks for face PAD under a unified linear-probing protocol on the four MCIO datasets, reporting intra-dataset and cross-dataset evaluation with ACER (\%), generalisation-gap analysis, and accuracy--compute trade-offs against PAD specialist baselines, i.e., the DNN DeepPixBiS and the ViT FSFM-FAS.
Remarkably, several foundation-model encoders achieve competitive PAD performance in the intra-dataset setting without any backbone fine-tuning, demonstrating that spoof-related cues are already encoded in their pretrained representations.
Despite recent studies on CLIP and DINO-based PAD systems, a broad and unified comparison across diverse foundation-model families and MLLM vision encoders remains limited~\cite{gonzalez2025foundation,feng2026benchmarking}.

% Our study extends, whose linear-probing experiments were limited to DINO and CLIP, 
% and reveals that the general-purpose representational capacity of large FMs for PAD remains largely unexplored relative to fine-tuned, spoof-specific alternatives~\cite{}.
% Our study extends~\cite{gonzalez2025foundation}, whose linear-probing experiments were limited to DINO and CLIP.
% Moreover, in recent work, such as a VFM PAD detector~\cite{feng2026benchmarking} (whereas fine-tuned SSL backbones with spoof-specific augmentation and patch losses approach cross-dataset performance of large VFMs under MICO-style training), we find capabilities of FMs' general purpose are unexplored.
Our main findings are threefold. 
% First, backbone scale and pretraining objective strongly shape intra-dataset separability: InternViT-6B reaches sub-percent ACER on several splits with a single trainable FC layer. 
% Second, cross-dataset transfer remains difficult for nearly every model, though InternViT-6B outperforms all other evaluated encoders; critically, low intra-dataset ACER — as observed for DeepPixBiS — is not a reliable indicator of robust cross-dataset performance. Third, accuracy--compute trade-offs matter: compact specialists remain competitive at a fraction of the inference cost of multi-billion-parameter backbones, while mid-sized SSL and vision--language encoders offer practical compromises when compute is constrained.
First, backbone scale and pretraining objective strongly affect intra-dataset detection accuracy.
InternViT-6B reaches sub-percent error on several dataset splits with only one trainable FC layer.
Second, cross-dataset transfer remains difficult, while intra-dataset error remains low for most FMs and DNNs.
Third, model scale alone does not guarantee improved transferability.
CLIP ViT-B/32 provides the strongest cross-dataset transfer–compute trade-off, achieving the lowest mean off-diagonal ACER among most evaluated probes while requiring substantially less computation.
Compact specialists such as DeepPixBiS remain competitive at a fraction of the inference cost of multi-billion-parameter backbones, though they exhibit higher variance across source-target dataset pairs.

% Backbone scale and pretraining objective strongly shape intra-dataset separability.
% Cross-dataset transfer remains difficult for nearly every model, although InternViT-6B outperforms the other evaluated encoders.
% Low intra-dataset ACER, as observed for DeepPixBiS, is not a reliable indicator of robust cross-dataset PAD performance.
The results indicate that multimodal and face-spoofing-specialised pretraining do not automatically yield the most transferable linear features, while mid-sized SSL and vision--language encoders offer practical compromises when compute is limited.
Future work includes intermediate-layer probing, multi-source protocols (LOO, LSD), and task-adaptive readouts beyond a single linear layer~\cite{ozgur2025foundpad,gonzalez2025foundation,feng2026benchmarking}.

\section*{Acknowledgment}
This research was funded by the European Union project CarMen (Grant Agreement No. 101168325).

%% file: content/figures/linear_probe_arch.tex
\begin{figure}[!t]
  \centering
  \resizebox{\columnwidth}{!}{%
  \begin{tikzpicture}[
    >=Stealth,
    font=\small,
    imgbox/.style={
      draw,
      rounded corners=2pt,
      minimum width=1.5cm,
      minimum height=1.5cm,
      fill=white,
      line width=0.6pt,
      align=center,
      inner sep=0pt,
    },
    vitblock/.style={
      draw,
      rounded corners=2pt,
      align=center,
      minimum width=2.6cm,
      minimum height=1.15cm,
      fill=white,
      line width=0.5pt,
      font=\scriptsize,
    },
    frozenbox/.style={
      draw,
      dashed,
      rounded corners=2pt,
      line width=0.6pt,
      fill=Gray,
      inner sep=7pt,
    },
    trainable/.style={
      draw,
      rounded corners=2pt,
      align=center,
      minimum height=9mm,
      line width=0.9pt,
      fill=white,
    },
    op/.style={
      draw,
      circle,
      minimum size=7mm,
      align=center,
      line width=0.6pt,
      fill=white,
    },
    io/.style={
      draw,
      rounded corners=2pt,
      align=center,
      minimum height=9mm,
      line width=0.6pt,
      fill=white,
    },
  ]
    % --- frozen ViT contents ---
    \node[anchor=north west, inner sep=0pt] (etitle) at (0, 0) {%
      \raisebox{-0.1ex}{\iconfrozen}\hspace{0.6mm}{\scriptsize\bfseries Frozen Foundation model}%
    };

    \node[vitblock, below=3mm of etitle.south west, anchor=north west] (vit)
      {Vision Transformer\\[-2pt]{\scriptsize $\times L$ layers}};

    \node[draw, rounded corners=1.5pt, fill=white, line width=0.5pt,
          minimum width=1.05cm, minimum height=7mm, align=center,
          anchor=west] (clsout) at ([xshift=3mm]vit.east)
      {CLS\\[-2pt]$h\!\in\!\mathbb{R}^{d}$};

    \coordinate (flowY) at (clsout);

    \begin{scope}[on background layer]
      \node[frozenbox, fit=(etitle)(vit)(clsout)] (encoder) {};
    \end{scope}

    \node[trainable, right=5mm of encoder, minimum width=1.25cm, anchor=west]
      (fc) at (encoder.east |- flowY)
      {Fully-Connected\\[-1pt]$d\!\to\!1$\\[-2pt]{\scriptsize trainable}};

    \node[anchor=south, inner sep=0pt] at ([yshift=2.2mm]fc.north) {\icontrainable};
    \node[op, right=4mm of fc] (sig) {$\sigma$};
    \node[io, right=4mm of sig] (out) {PAD score\\[-1pt]$p$};

    \node[imgbox, anchor=east] (input) at ([xshift=-5mm]encoder.west) {\iconface};
    \node[font=\scriptsize, below=0.8mm of input.south] {Face image $x$};

    \draw[->] (input.east) -- (encoder.west);
    \draw[->] (clsout.east) -- (fc.west);
    \draw[->] (fc) -- (sig);
    \draw[->] (sig) -- (out);
  \end{tikzpicture}%
  }
  \caption{
  Overview of the linear probing for binary detection:
  A frozen pre-trained ViT vision encoder maps the face image to a CLS embedding~$h$ via $L$ Vision Transformer (ViT) layers. 
  In probing, only the fully-connected layer (FC) is trainable cf.\ Table~\ref{tab:indomain-acer}), and has the input shape of the CLS embedding $d$ mapping to $1$.
  The probability of a sample being bonafide is $p=\sigma(z)$, where $z$ is the linear output (logit) of the FC, and the sigmoid function $\sigma(\cdot)$ converts the logit to a probability for binary classification.
  The target label is 1 for bonafide samples and 0 for attacked samples.
  }
  \label{fig:linear-probe-arch}
\end{figure}

%% file: content/tables/encoder_taxonomy.tex
\begin{table}[t]
  \centering
  \caption{
  Frozen model backbone zoo by family for linear probing.
  Within the family, the models only differ in parameter size (in Billions).
  }
  \label{tab:encoder-taxonomy}
  \scriptsize
  \setlength{\tabcolsep}{1.5pt}
  \resizebox{\columnwidth}{!}{%
  \begin{tabular}{@{}llrll@{}}
    \toprule
    Family & Variants & \makecell[c]{Params\\(B)} & Supervision & Pretraining objective \\
    \midrule
    ConvNeXt-V2 & base, huge       & 0.09, 0.66   & Supervised    & ImageNet classification \\
    DINOv2      & base, giant, w/reg. & 0.09--1.14 & SSL       & Self-distillation       \\
    DINOv3      & base, giant      & 0.09, 0.84   & SSL           & Self-distillation       \\
    BEiT        & base, large      & 0.09, 0.30   & SSL           & Masked image modeling   \\
    ViT-MAE     & base, huge       & 0.09, 0.63   & SSL           & Masked autoencoding     \\
    ViT-MSN     & base, large      & 0.09, 0.30   & SSL           & Masked neighbours       \\
    CLIP        & B/32, L/14, H/14 & 0.09--0.63   & Vision--lang. & Img--txt contrastive    \\
    SigLIP      & base, large      & 0.09, 0.32   & Vision--lang. & Sigmoid contrastive     \\
    EVA-CLIP    & 8B               & 8.22         & Vision--lang. & Img--txt contrastive    \\
    Qwen2.5-VL  & 3B, 7B           & 0.67, 0.68   & Multimodal    & Img--txt contrastive  \\
    InternViT   & 300M, 6B         & 0.30, 5.54   & Multimodal    & VLM vision tower        \\
    \bottomrule
  \end{tabular}
  }
  \vspace{0.5mm}
  {\scriptsize
    SSL: self-supervised (no manual class labels).
    Supervised: ImageNet-1K classification (ConvNeXt-V2 CNN).
    Vision--lang.: image--text pair pretraining.
    Qwen2.5-VL variants extract the frozen ViT from 3B/7B VLM checkpoints; both towers are $\sim$0.67\,B.
    Native input 224\,px except SigLIP-large (256\,px) and InternViT (448\,px).}
\end{table}

%% file: content/figures/mcio_samples.tex
\begin{figure}[t]
  \centering
  \includegraphics[width=\columnwidth]{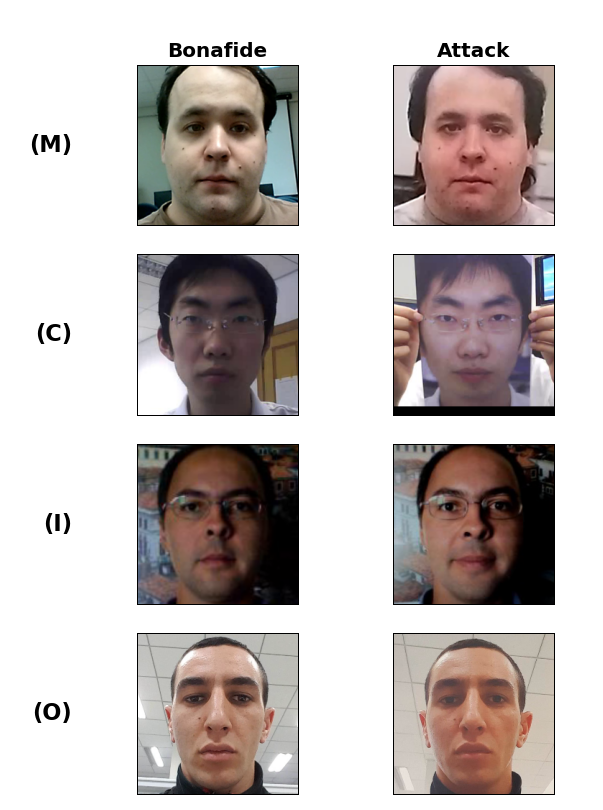}
  \caption{
  Examples face crops with the MTCNN detector \cite{MTCNN}. 
  {
  \scriptsize
  Left column: bonafide. Right column: presentation attack (type: print). 
  Rows - datasets: MSU-MFSD (M), CASIA-FASD (C), Replay-Attack (I), Oulu-NPU (O).
  }
  }
  \label{fig:mcio-samples}
\end{figure}

%% file: content/tables/indomain_acer.tex
\begin{table*}[htbp]
  \centering
  \caption{
  Result: intra-dataset using the ACER (\%) metric. The threshold $\tau$ (cf. \cref{eq:acer})  is calculated on the dev split and applied to the test split.
  The last two columns show the avg.\ is the unweighted mean and std.\ the standard deviation across the four MCIO corpora. 
  The parameter sizes are marked with frozen \iconfrozen\hspace{0.5mm} (in Billion - $10^9$) for the foundation model and trainable \icontrainable\hspace{0.5mm} (in Kilo - $10^3$) for the fully connected layer.
  The PAD specialist baselines (DeepPixBiS and FSFM-FAS) are fully trainable.
  {\scriptsize 
  The feature denotes how each frozen encoder is pooled to one vector
(CLS token, pooling: post-layer-norm CLS, global average pooling, contrastive embedding, or mean-pooled patches)
  Datasets: M=MSU-MFSD, C=CASIA-FASD, I=Replay-Attack, O=OULU-NPU.
  Bold: lowest ACER per column. 
  }
  }
  \label{tab:indomain-acer}
  \footnotesize
  \resizebox{0.9\linewidth}{!}{%
  \begin{tabular}{lccccccccc}
    \toprule
    \multirow{2}{*}{Model} & \multirow{2}{*}{Feature} & \multicolumn{2}{c}{Parameters} & \multicolumn{6}{c}{Datasets} \\
    \cmidrule(lr){3-4} \cmidrule(lr){5-10}

    &  &{\iconfrozen$\times 10^9$}\hspace{0.5mm} & {\icontrainable$\times 10^3$}\hspace{0.5mm} & M & C & I & O & Avg. & Std. \\
    \midrule
    \multicolumn{10}{c}{Model Zoo} \\
    \midrule  % Added horizontal line
    ConvNeXt-V2-base~\cite{woo2023convnext}                   & GAP+LN & 0.09 & 1.0 & 24.7 & 13.7 & 13.7 & 33.5 & 21.4 & 8.3 \\
    ConvNeXt-V2-huge~\cite{woo2023convnext}                   & GAP+LN & 0.66 & 2.8 & 23.7 & 8.4 & 14.4 & 26.5 & 18.2 & 7.2 \\
    DINOv2-base~\cite{oquab2023dinov2}                        & CLS & 0.09 & 0.8 & 20.2 & 2.7 & 9.8 & 20.2 & 13.2 & 7.4 \\
    % DINOv2-large                                            & CLS & 0.30 & 1.0 & 15.1 & 9.8 & 7.5 & 25.8 & 14.5 & 7.1 \\
    DINOv2-giant~\cite{oquab2023dinov2}                       & CLS & 1.14 & 1.5 & 10.8 & 5.5 & 3.6 & 13.8 & 8.4 & 4.1 \\
    DINOv2-with-registers-base~\cite{darcet2024registers}     & CLS & 0.09 & 0.8 & 16.9 & 7.6 & 4.9 & 28.9 & 14.6 & 9.4 \\
    DINOv2-with-registers-giant~\cite{darcet2024registers}    & CLS & 1.14 & 1.5 & 9.5 & 5.0 & 8.4 & 11.4 & 8.6 & 2.3 \\
    DINOv3-base~\cite{simeoni2025dinov3}                      & CLS & 0.09 & 0.8 & 12.8 & 3.8 & 6.1 & 14.9 & 9.4 & 4.6 \\
    DINOv3-giant~\cite{simeoni2025dinov3}                     & CLS & 0.84 & 1.3 & 8.8 & \bestcell{0.6} & 7.9 & 5.5 & 5.7 & 3.2 \\
    BEiT-base~\cite{bao2021beit}                              & Post-LN CLS & 0.09 & 0.8 & 12.7 & 6.6 & 8.6 & 8.9 & 9.2 & 2.2 \\
    BEiT-large~\cite{bao2021beit}                             & Post-LN CLS & 0.30 & 1.0 & 14.5 & 9.0 & 5.0 & 7.9 & 9.1 & 3.4 \\
    % ConvNext-Base-22k                                       & & 0.09 & 1.0 & 19.1 & 10.2 & 6.0 & 18.4 & 13.4 & 5.5 \\
    ViT-MSN-base~\cite{assran2022masked}                      & CLS & 0.09 & 0.8 & 47.9 & 51.7 & 50.6 & 36.1 & 46.5 & 6.2 \\
    ViT-MSN-large~\cite{assran2022masked}                     & CLS & 0.30 & 1.0 & 49.7 & 51.0 & 42.7 & 37.5 & 45.2 & 5.5 \\
    ViT-MAE-base~\cite{he2022masked}                          & CLS & 0.09 & 0.8 & 20.6 & 14.3 & 7.4 & 31.6 & 18.5 & 8.9 \\
    ViT-MAE-huge~\cite{he2022masked}                          & CLS & 0.63 & 1.3 & 16.4 & 23.4 & 2.1 & 24.8 & 16.7 & 9.0 \\
    SigLIP-base~\cite{zhai2023sigmoid}                        & Post-LN CLS & 0.09 & 0.8 & 22.4 & 14.3 & 16.5 & 36.8 & 22.5 & 8.8 \\
    SigLIP-large~\cite{zhai2023sigmoid}                       & Post-LN CLS & 0.32 & 1.0 & 27.3 & 9.5 & 24.4 & 15.9 & 19.3 & 7.0 \\
    % siglip-so400m-patch14-384                               & & 0.43 & 1.2 & 29.9 & 6.1 & 22.9 & 18.2 & 19.3 & 8.7 \\
    CLIP ViT-B/32~\cite{radford2021learning}                  & Post-LN CLS & 0.09 & 0.8 & 21.8 & 4.4 & 18.6 & 5.7 & 12.6 & 7.7 \\
    CLIP ViT-L/14~\cite{radford2021learning}                  & Post-LN CLS & 0.30 & 1.0 & 19.0 & 6.9 & 9.5  & 7.1 & 10.6 & 4.9 \\
    CLIP-ViT-H/14~\cite{radford2021learning}                  & Post-LN CLS & 0.63 & 1.3 & 15.2 & 3.1 & 16.4 & 7.3 & 10.5 & 5.5 \\
    EVA-CLIP-8B~\cite{sun2023eva}                             & Contrastive-embedding & 8.22 & 1.3 & 12.2 & 2.1 & 15.3 & 11.2 & 10.2 & 4.9 \\
    Qwen2.5-VL-3B-ViT~\cite{bai2023qwen}                      & Mean pool & 0.67 & 2.0 & 19.6 & 6.5 & 21.3 & 17.1 & 16.1 & 5.8 \\
    Qwen2.5-VL-7B-ViT~\cite{bai2023qwen}                      & Mean pool & 0.68 & 3.6 & 17.2 & 6.1 & 27.6 & 18.1 & 17.2 & 7.6 \\
    InternViT-300M~\cite{chen2024internvl}                    & CLS & 0.30 & 1.0 & 22.2 & 34.5 & 16.1 & 22.7 & 23.9 & 6.7 \\
    \htablegrayrow{InternViT-6B  \cite{chen2024internvl}      & CLS & 5.54 & 3.2 & 4.4 & 1.8 & 0.1 & \bestcell{0.3} & \bestcell{1.6} & \bestcell{1.7}} \\
    \midrule
    \multicolumn{10}{c}{Baseline} \\
    \midrule  % Added horizontal line
    InternVL3-14B (MLLM; 0-Shot) \cite{chen2024internvl}            & --- & 14.0 & --- & 38.4 & 14.4 & 48.8 & 39.6 & 35.3 & 12.7 \\
    FSFM-FAS (MAE~\cite{he2022masked}; ViT-B16)~\cite{wang2025fsfm} & --- & --- & $86\times 10^3$ & \bestcell{2.8} & 1.6 & 0.2 & 12.0 & 4.2 & 4.6 \\
    \htablegrayrow{DeepPixBiS (DNN)~\cite{george2019deep}           & --- & --- & $3\times 10^3$ & 7.9 & 2.0 & \bestcell{0.0} & 3.8 & 3.4 & 2.9} \\
    \bottomrule
  \end{tabular}
  }
\end{table*}

%% file: content/tables/cross_dataset_representative.tex
\begin{table}[t]
  \centering
  \caption{
  Cross-domain ACER (\%) for InternViT-6B and DeepPixBiS.
  The highest transferability is observed for InternViT-6B when trained on CASIA-FASD (C).
  Diagonal cells are \colorbox{yellow!30}{intra-dataset (ID)}.
  Off-diagonal cells expose \colorbox{green!30}{cross-dataset (CD)} transfer.
  Full matrices for all benchmarked models are in Appendix~\ref{app:cross-dataset-generalizability}.
  }
  \label{tab:cross-dataset-representative}
  \scriptsize
  \setlength{\tabcolsep}{2pt}
  \resizebox{0.9\columnwidth}{!}{%
  \begin{tabular}{@{}lccccccc@{}}
    \toprule
    \multirow{2}{*}{Model} & \multirow{2}{*}{Train} & \multicolumn{6}{c}{Test in ACER (\%)} \\
    \cmidrule(lr){3-8}
    & & M & C & I & O & Avg. & Std. \\
    \midrule
    InternViT-6B & M & \cellcolor{yellow!30}{4.42} & \cellcolor{green!20}{16.64} & \cellcolor{green!20}{45.29} & \cellcolor{green!20}{14.65} & 20.25 & 15.18 \\
     & C & \cellcolor{green!20}{28.07} & \cellcolor{yellow!30}{1.81} & \cellcolor{green!20}{21.26} & \cellcolor{green!20}{25.86} & \bestcell{19.25} & \bestcell{10.36} \\
     & I & \cellcolor{green!20}{28.54} & \cellcolor{green!20}{25.26} & \cellcolor{yellow!30}{0.06} & \cellcolor{green!20}{50.33} & 26.05 & 17.83 \\
     & O & \cellcolor{green!20}{48.50} & \cellcolor{green!20}{38.89} & \cellcolor{green!20}{40.37} & \cellcolor{yellow!30}{0.27} & 32.01 & 18.68 \\
    \midrule
    DeepPixBiS & M & \cellcolor{yellow!30}{7.89} & \cellcolor{green!20}{47.05} & \cellcolor{green!20}{49.65} & \cellcolor{green!20}{36.46} & 35.26 & 16.56 \\
     & C & \cellcolor{green!20}{33.74} & \cellcolor{yellow!30}{2.01} & \cellcolor{green!20}{30.74} & \cellcolor{green!20}{30.20} & \bestcell{24.17} & \bestcell{12.87} \\
     & I & \cellcolor{green!20}{47.36} & \cellcolor{green!20}{61.34} & \cellcolor{yellow!30}{0.00} & \cellcolor{green!20}{50.94} & 39.91 & 23.61 \\
     & O & \cellcolor{green!20}{46.55} & \cellcolor{green!20}{38.36} & \cellcolor{green!20}{45.10} & \cellcolor{yellow!30}{3.84} & 33.46 & 17.38 \\
    \bottomrule
  \end{tabular}
  }
\end{table}

%% file: content/tables/cross_dataset_summary.tex
\begin{table}[t]
  \centering
  \caption{
      Summary in the ACER (\%) metric.
      Average over intra-dataset and cross-dataset (\cref{eq:mean-acer-all}).
      \colorbox{yellow!30}{Intra-dataset (ID)}: mean of the four intra-dataset  evaluation per model.
      \colorbox{green!30}{Cross-dataset (CD)} (\cref{eq:mean-acer-offdiag}): mean of the twelve cross-datasets (transferability). 
  Gap $=$ CD $-$ ID. 
  {
  \scriptsize
    \textbf{Bold}: lowest value per column. All values in percent.
  }
  }
  \label{tab:cross-dataset-summary}
  \scriptsize
  \setlength{\tabcolsep}{2pt}
  \resizebox{0.8\columnwidth}{!}{%
  \begin{tabular}{@{}lrrrr@{}}
    \toprule
    Model & All & \cellcolor{yellow!30}{ID} & \cellcolor{green!30}{CD} & Gap \\
    \midrule
    InternViT-6B & \bestcell{24.4} & \bestcell{1.6} & 32.0 & 30.3 \\
    CLIP ViT-B/32 & 26.8 & 12.6 & \bestcell{31.6} & 18.9 \\
    CLIP ViT-L/14 & 28.2 & 10.6 & 34.0 & 23.4 \\
    FSFM-FAS & 29.0 & 4.2 & 37.3 & 33.1 \\
    BEiT-large & 29.3 & 9.1 & 36.0 & 26.9 \\
    EVA-CLIP-8B & 30.4 & 10.2 & 37.1 & 26.9 \\
    CLIP ViT-H/14 & 31.6 & 10.5 & 38.7 & 28.2 \\
    DINOv2-giant & 31.9 & 8.4 & 39.7 & 31.3 \\
    DINOv2-registers-giant & 31.9 & 8.6 & 39.7 & 31.1 \\
    DINOv3-giant & 32.7 & 5.7 & 41.8 & 36.1 \\
    DINOv3-base & 33.2 & 9.4 & 41.1 & 31.7 \\
    DeepPixBiS & 33.2 & 3.4 & 43.1 & 39.7 \\
    DINOv2-registers-base & 34.4 & 14.6 & 40.9 & 26.4 \\
    BEiT-base & 35.0 & 9.2 & 43.7 & 34.5 \\
    Qwen2.5-VL-7B-ViT & 34.8 & 17.2 & 40.7 & 23.4 \\
    DINOv2-base & 35.5 & 13.2 & 42.9 & 29.7 \\
    ViT-MAE-huge & 36.2 & 16.7 & 42.7 & 26.1 \\
    SigLIP-large & 36.3 & 19.3 & 42.0 & 22.8 \\
    Qwen2.5-VL-3B-ViT & 36.5 & 16.1 & 43.2 & 27.1 \\
    SigLIP-base & 39.5 & 22.5 & 45.1 & 22.6 \\
    ConvNeXt-V2-huge & 40.5 & 18.2 & 48.0 & 29.7 \\
    ConvNeXt-V2-base & 41.3 & 21.4 & 47.9 & 26.5 \\
    ViT-MAE-base & 41.7 & 18.5 & 49.4 & 31.0 \\
    InternViT-300M & 41.8 & 23.9 & 47.7 & 23.9 \\
    ViT-MSN-large & 46.8 & 45.2 & 47.3 & \bestcell{2.0} \\
    ViT-MSN-base & 53.6 & 46.5 & 56.0 & 9.4 \\
    \bottomrule
  \end{tabular}
  }
\end{table}

%% file: content/supplementary.tex
\appendix

\section{ACER metric definitions}
\label{app:metrics}

We follow ISO/IEC~30107-3~\cite{ISO301073} and standard PAD practice~\cite{DBLP:conf/icmcs/LiuCDLZX22,Fang_2022_WACV,fang2024face}.
Labels are bonafide ($y{=}1$) and attack ($y{=}0$).
Let $\mathcal{P}_{\mathrm{bon}}$ and $\mathcal{P}_{\mathrm{att}}$ denote bonafide and attack test samples, and $\hat{y}_\tau(x)\in\{0,1\}$ the hard decision at threshold~$\tau$.
The \emph{Attack Presentation Classification Error Rate} (APCER) and \emph{Bonafide Presentation Classification Error Rate} (BPCER) are
\begin{align}
  \mathrm{APCER}(\tau)
  &= \frac{1}{|\mathcal{P}_{\mathrm{att}}|}
     \sum_{x \in \mathcal{P}_{\mathrm{att}}} \mathbb{I}\!\left[\hat{y}_\tau(x)=1\right],
  \label{eq:apcer} \\
  \mathrm{BPCER}(\tau)
  &= \frac{1}{|\mathcal{P}_{\mathrm{bon}}|}
     \sum_{x \in \mathcal{P}_{\mathrm{bon}}} \mathbb{I}\!\left[\hat{y}_\tau(x)=0\right],
  \label{eq:bpcer}
\end{align}
where $\mathbb{I}[\cdot]$ is the indicator function.
ACER is $\mathrm{ACER}(\tau)=\tfrac{1}{2}(\mathrm{APCER}(\tau)+\mathrm{BPCER}(\tau))$ (Eq.~\eqref{eq:acer} in the main paper).
The operating threshold is selected on the development split by minimizing $|\mathrm{APCER}(\tau)-\mathrm{BPCER}(\tau)|$ (EER criterion) and applied unchanged to the test split.

\section{InternViT encoder details}
\label{app:internvit}

InternViT is the vision tower of the InternVL open-source multimodal family~\cite{chen2024internvl,chen2024internvl2}.
We use the standalone V2.5 checkpoints InternViT-300M and InternViT-6B, released without the language model or MLP connector (\texttt{InternVisionModel} in Hugging Face).
Both are patch-based ViTs with $14{\times}14$ conv patch embedding, $448{\times}448$ input resolution, GELU feed-forward blocks, and pre-norm LayerNorm.
InternViT-300M has 24 transformer layers, hidden size $d{=}1024$, 16 attention heads, and MLP width 4096 (${\sim}304$\,M parameters).
InternViT-6B scales to 45 layers, $d{=}3200$, 25 heads, MLP width 12800, and QK normalization (${\sim}5.5$\,B parameters).
We take the CLS token from the final layer as the global face embedding.
The full InternVL stack (pixel unshuffle, dynamic tiling, MLP projector, and LLM) is not used at inference.

Pretraining follows the InternVL~2.5 recipe~\cite{chen2024internvl2} in a ViT--MLP--LLM stack.
In Stage~1, only the MLP connector is trained while the vision encoder and language model remain frozen.
In optional Stage~1.5 (ViT incremental learning), the vision encoder and MLP are updated jointly using next-token prediction loss on multimodal data.
In Stage~2 (instruction tuning), the full model is trained end-to-end on multimodal instruction datasets.
The Hugging Face checkpoints retain weights from this joint vision--text pretraining.
In our PAD experiments, the encoder is kept frozen, and only the linear probe is trained.

\section{PAD dataset sample counts}
\label{app:dataset-counts}

\begin{figure*}[htbp]
    \centering
    \includegraphics[width=\linewidth]{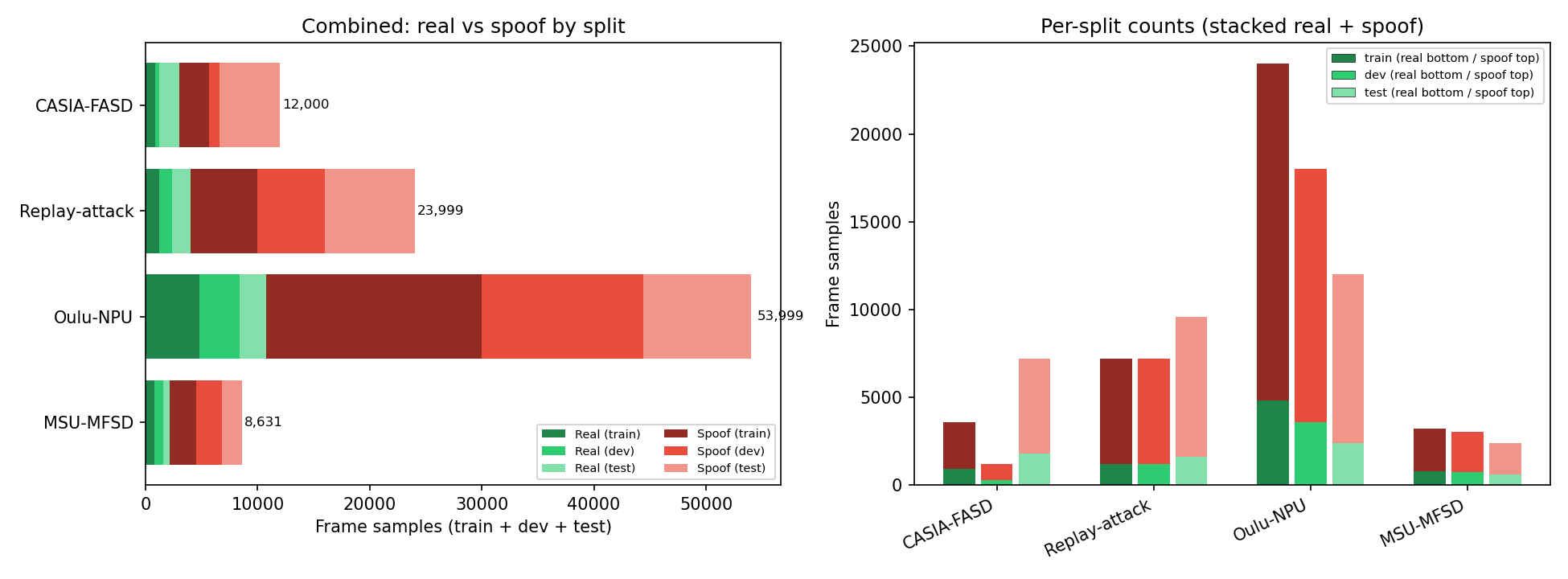}
    \caption{Combined summary}
    \label{fig:app-dataset-counts-combined}
  \caption{
  PAD dataset sample counts per protocol split (train, test, dev). Green: bonafide (real); red: attack (spoof). 
  Counts are post-preprocess frame manifests (successful MTCNN extractions). Up to 20 frames per video.
  }
  \label{fig:app-dataset-counts}
\end{figure*}

Frame-level sample counts after face detection and cropping.
We report the four MCIO datasets in~\cref{fig:app-dataset-counts}.

% \begin{figure*}[htbp]
%   \centering
%   \begin{subfigure}[t]{0.48\linewidth}
%     \centering
%     \includegraphics[width=\linewidth]{pad/datasets/pad_dataset_sample_counts_train.png}
%     \caption{Train split}
%     \label{fig:app-dataset-counts-train}
%   \end{subfigure}
%   \hfill
%   \begin{subfigure}[t]{0.48\linewidth}
%     \centering
%     \includegraphics[width=\linewidth]{pad/datasets/pad_dataset_sample_counts_dev.png}
%     \caption{Dev split}
%     \label{fig:app-dataset-counts-dev}
%   \end{subfigure}
%   \vspace{0.5em}
%   \begin{subfigure}[t]{0.48\linewidth}
%     \centering
%     \includegraphics[width=\linewidth]{pad/datasets/pad_dataset_sample_counts_test.png}
%     \caption{Test split}
%     \label{fig:app-dataset-counts-test}
%   \end{subfigure}
%   \hfill
%   \begin{subfigure}[t]{0.48\linewidth}
%     \centering
%     \includegraphics[width=\linewidth]{pad/datasets/pad_dataset_sample_counts.png}
%     \caption{Combined summary}
%     \label{fig:app-dataset-counts-combined}
%   \end{subfigure}
%   \caption{
%   PAD dataset sample counts per protocol split (train, test, dev). Green: bonafide (real); red: attack (spoof). 
%   Counts are post-preprocess frame manifests (successful MTCNN extractions). Up to 20 frames per video.
%   }
%   \label{fig:app-dataset-counts}
% \end{figure*}

% BEGIN AUTO-GENERATED CROSS-DATASET APPENDIX

\section{Cross-dataset generalizability}
\label{app:cross-dataset-generalizability}

Full MCIO cross-dataset test ACER results for all benchmarked models.
Each block lists one model; rows indicate the training corpus (M=MSU-MFSD, C=CASIA-FASD, I=Replay-Attack, O=Oulu-NPU) and columns the test corpus.
Bold entries mark the lowest value in each column within the table (ACER in M--O; Avg./Std.\ as defined below; lower is better).
Matrix aggregates ${\mathrm{ACER}}_{\mathrm{all}}$ and ${\mathrm{ACER}}_{\mathrm{cross}}$ are defined in Eqs.~\eqref{eq:mean-acer-all}--\eqref{eq:mean-acer-offdiag} (summarized in Table~\ref{tab:cross-dataset-summary}).
Avg.\ and Std.\ are the unweighted mean and population standard deviation across the four test corpora for each training source.
All values are in percent (\%); lower is better.
Results can be found in the following~\cref{tab:cross-dataset-acer,tab:cross-dataset-acer-cont1,tab:cross-dataset-acer-cont2,tab:cross-dataset-acer-cont3}.

\section{
GFLOPs measurement on DeepPixBis PAD detector on two image sizes (224 vs 448 pixels).
}
We compare two profiling libraries on three models in~\cref{tab:gflops_comparison}:
Torch\footnote{\texttt{from torch.utils.flop\_counter import FlopCounterMode}} and
Thop\footnote{\url{https://github.com/ultralytics/thop}}.
The difference between the GLOPs measurements of both libraries is small. 
\begin{table}[H]
\caption{
GFLOPs comparison between Torch and THOP library.
The number of GFLOPs increases four times by doubling the image size on DeepPixBis.
The other models are rigid to the input size.
}
\label{tab:gflops_comparison}
\centering
\resizebox{1\linewidth}{!}{%
\begin{tabular}{lrrrrl}
\toprule
Model & Size & Torch GFLOPs & THOP GFLOPs & Diff (\%) & Status \\
\midrule
DeepPixBiS   & 224 & 9.169      & 9.304      & 1.47  & ok   \\
DeepPixBiS   & 448 & 36.676     & 37.215     & 1.47  & ok   \\
InternViT-6B & 224 & ---        & ---        & ---   & fail \\
InternViT-6B & 448 & 11944.694  & 11339.534  & -5.07 & ok   \\
CLIP-B/32    & 224 & 8.725      & 8.732      & 0.09  & ok   \\
CLIP-B/32    & 448 & ---        & ---        & ---   & fail \\
\bottomrule
\end{tabular}
}
\end{table}

% \FloatBarrier
\section{CLS embedding t-SNE and UMAP visualizations}
\label{app:tsne-cls}

Each panel shows a two-dimensional t-SNE and UMAP projection of frozen CLS embeddings from the MCIO test split (MSU-MFSD, CASIA-FASD, Replay-Attack, Oulu-NPU), with all corpora pooled in one embedding space.
t-SNE: Features are reduced with PCA ($d{=}64$) before t-SNE (perplexity $30$, learning rate $200$, $1000$ iterations).
UMAP: Features are reduced with PCA ($d{=}64$) before UMAP ($n_{\mathrm{neighbors}}{=}20$, $\mathrm{min\_dist}{=}0.1$, cosine metric).
Dataset-colored plots use distinct markers and hues per corpus; attack points within each corpus share the corpus color, and bonafide samples are shown in red.

\begin{figure*}[p]
  \centering
  \begin{subfigure}[t]{0.48\linewidth}
    \centering
    \includegraphics[width=\linewidth]{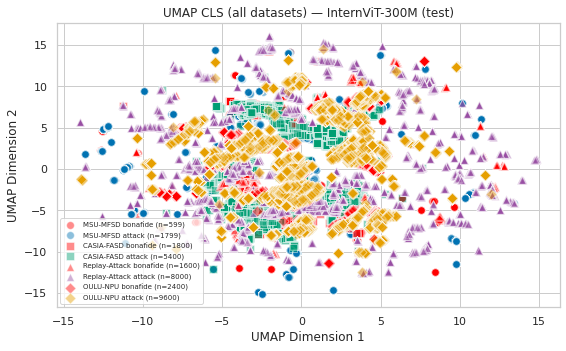}
    \caption{InternViT-300M}
    \label{fig:app-umap-internvit-300m-448px-v2-5-by-dataset}
  \end{subfigure}  \hfill
  \begin{subfigure}[t]{0.48\linewidth}
    \centering
    \includegraphics[width=\linewidth]{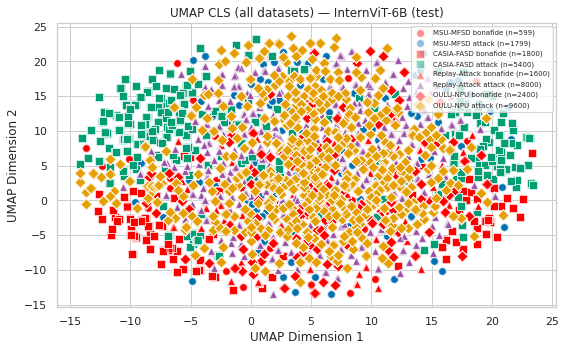}
    \caption{InternViT-6B}
    \label{fig:app-umap-internvit-6b-448px-v2-5-by-dataset}
  \end{subfigure}
  \caption{t-SNE are in \cref{fig:tsne-main}. 
  UMAP of frozen CLS embeddings (MCIO test samples, all corpora pooled; colored by dataset).}
  \label{fig:app-umap-by-dataset-5}
\end{figure*}

\begin{figure*}[p]
  \centering
  \begin{subfigure}[t]{0.48\linewidth}
    \centering
    \includegraphics[width=\linewidth]{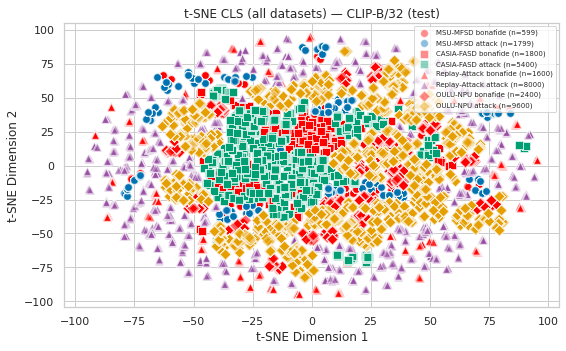}
    \caption{CLIP ViT-B/32}
    \label{fig:app-tsne-openai-clip-vit-base-patch32-by-dataset}
  \end{subfigure}  \hfill
  \begin{subfigure}[t]{0.48\linewidth}
    \centering
    \includegraphics[width=\linewidth]{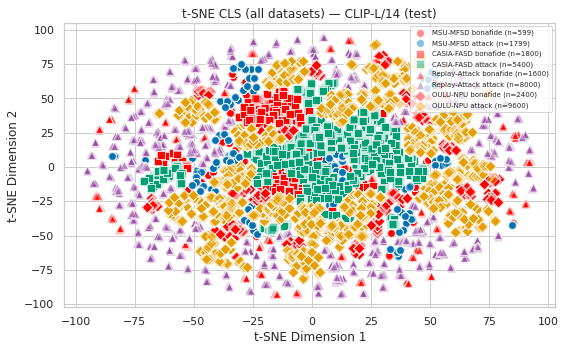}
    \caption{CLIP ViT-L/14}
    \label{fig:app-tsne-openai-clip-vit-large-patch14-by-dataset}
  \end{subfigure}
    \vspace{0.5em}
  \begin{subfigure}[t]{0.48\linewidth}
    \centering
    \includegraphics[width=\linewidth]{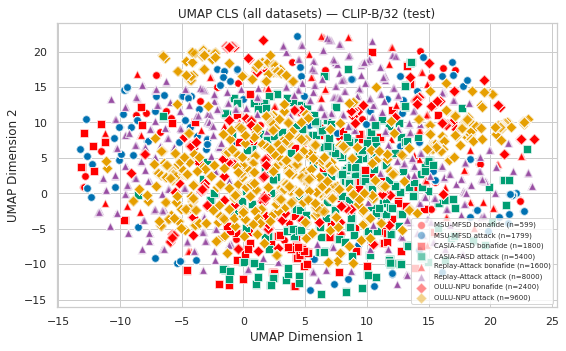}
    \caption{CLIP ViT-B/32}
    \label{fig:app-umap-openai-clip-vit-base-patch32-by-dataset}
  \end{subfigure}  \hfill
  \begin{subfigure}[t]{0.48\linewidth}
    \centering
    \includegraphics[width=\linewidth]{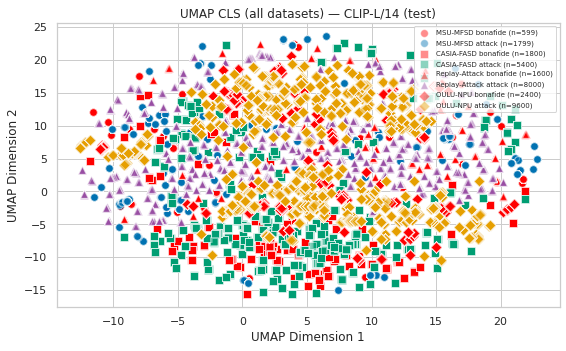}
    \caption{CLIP ViT-L/14}
    \label{fig:app-umap-openai-clip-vit-large-patch14-by-dataset}
  \end{subfigure}
  \caption{Top: t-SNE of frozen CLS embeddings (MCIO test samples, all corpora pooled; colored by dataset). 
  Bottom: corresponding UMAP.
  }
  \label{fig:app-tsne-by-dataset-1}
\end{figure*}

Plot for InternVITs and CLIP-B/32 can be found in~\cref{fig:app-umap-by-dataset-5,fig:app-tsne-by-dataset-1}.

\begin{table*}[htbp]
  \centering
  \caption{Cross-dataset test ACER (\%) for MCIO leave-one-out training. Rows: training corpus; columns: test corpus. Bold: lowest ACER per column within the table. Avg./Std.: mean and population std.\ over the four test corpora per train source.}
  \label{tab:cross-dataset-acer}
  \footnotesize
  \resizebox{0.7\linewidth}{!}{%
\begin{tabular}{llcccccc}
\toprule
\multirow{2}{*}{Model} & \multirow{2}{*}{Train} & \multicolumn{6}{c}{Test} \\
\cmidrule(lr){3-8}
& & M & C & I & O & Avg. & Std. \\
\midrule
CLIP ViT-B/32 & M & 21.81 & 48.25 & 48.11 & 21.84 & 35.00 & 13.18 \\
 & C & 27.35 & 4.39 & 34.77 & 18.68 & 21.30 & 11.30 \\
 & I & 26.61 & 36.70 & 18.62 & 27.44 & 27.34 & \bestcell{6.41} \\
 & O & 25.69 & 24.02 & 39.30 & \bestcell{5.74} & 23.69 & 11.94 \\
\midrule
CLIP ViT-L/14 & M & 18.99 & 44.13 & 35.04 & 43.86 & 35.50 & 10.21 \\
 & C & 27.30 & 6.94 & 24.75 & 19.16 & \bestcell{19.54} & 7.85 \\
 & I & 29.21 & 35.45 & 9.54 & 46.28 & 30.12 & 13.36 \\
 & O & 19.60 & 36.77 & 46.96 & 7.11 & 27.61 & 15.35 \\
\midrule
CLIP ViT-H/14 & M & 15.23 & 35.42 & 41.25 & 36.71 & 32.15 & 10.01 \\
 & C & 30.52 & 3.09 & 33.98 & 41.34 & 27.23 & 14.47 \\
 & I & 35.18 & 31.64 & 16.38 & 43.60 & 31.70 & 9.86 \\
 & O & 47.66 & 39.80 & 46.97 & 7.29 & 35.43 & 16.54 \\
\midrule
DINOv2-base & M & 20.16 & 36.82 & 46.41 & 37.62 & 35.25 & 9.49 \\
 & C & 43.05 & \bestcell{2.67} & 43.31 & 49.26 & 34.57 & 18.59 \\
 & I & 35.78 & 32.99 & 9.76 & 44.19 & 30.68 & 12.76 \\
 & O & 52.27 & 42.66 & 50.61 & 20.25 & 41.45 & 12.77 \\
\midrule
DINOv2-giant & M & 10.84 & 32.76 & 45.68 & 33.38 & 30.67 & 12.55 \\
 & C & 40.83 & 5.46 & 35.99 & 33.05 & 28.83 & 13.78 \\
 & I & 45.33 & 25.59 & \bestcell{3.59} & 47.40 & 30.48 & 17.70 \\
 & O & 48.62 & 38.78 & 48.91 & 13.78 & 37.52 & 14.30 \\
\midrule
DINOv2-with-registers-base & M & 16.90 & 27.07 & 38.44 & 39.69 & 30.53 & 9.28 \\
 & C & 43.17 & 7.63 & 43.79 & 46.55 & 35.29 & 16.02 \\
 & I & 38.76 & 31.82 & 4.86 & 50.51 & 31.49 & 16.76 \\
 & O & 48.60 & 35.18 & 47.81 & 28.91 & 40.12 & 8.38 \\
\midrule
DINOv2-with-registers-giant & M & \bestcell{9.51} & 35.44 & 41.39 & 39.11 & 31.37 & 12.80 \\
 & C & 27.08 & 5.02 & 39.52 & 35.02 & 26.66 & 13.26 \\
 & I & 32.71 & 46.29 & 8.39 & 48.22 & 33.90 & 15.90 \\
 & O & 44.37 & 39.89 & 47.15 & 11.40 & 35.70 & 14.27 \\
\bottomrule
\end{tabular}
  }
\end{table*}

\begin{table*}[htbp]
  \centering
  \caption{Cross-dataset test ACER (\%) --- continued.}
  \label{tab:cross-dataset-acer-cont1}
  \footnotesize
  \resizebox{0.7\linewidth}{!}{%
\begin{tabular}{llcccccc}
\toprule
\multirow{2}{*}{Model} & \multirow{2}{*}{Train} & \multicolumn{6}{c}{Test} \\
\cmidrule(lr){3-8}
& & M & C & I & O & Avg. & Std. \\
\midrule
DINOv3-base & M & 12.79 & 41.07 & 44.58 & 30.16 & 32.15 & 12.38 \\
 & C & 44.89 & 3.81 & 38.42 & 38.41 & 31.38 & 16.13 \\
 & I & 49.72 & 47.94 & 6.08 & 49.98 & 38.43 & 18.70 \\
 & O & 30.43 & 39.59 & 38.17 & 14.89 & 30.77 & 9.81 \\
\midrule
DINOv3-giant & M & \bestcell{8.76} & 48.94 & 43.85 & 24.07 & 31.40 & 16.04 \\
 & C & 40.61 & \bestcell{0.58} & 45.17 & 33.09 & 29.86 & 17.45 \\
 & I & 36.03 & 48.92 & 7.94 & 48.20 & 35.27 & 16.59 \\
 & O & 35.56 & 46.71 & 50.00 & \bestcell{5.51} & 34.44 & 17.54 \\
\midrule
BEiT-base & M & 12.73 & 42.93 & 47.01 & 39.84 & 35.63 & 13.46 \\
 & C & 47.22 & 6.57 & 39.12 & 49.82 & 35.68 & 17.26 \\
 & I & 35.76 & 42.86 & 8.56 & 48.80 & 33.99 & 15.39 \\
 & O & 47.10 & 35.64 & 47.79 & 8.91 & 34.86 & 15.74 \\
\midrule
BEiT-large & M & 14.49 & 25.16 & 37.01 & 20.79 & \bestcell{24.36} & \bestcell{8.23} \\
 & C & 44.27 & 9.05 & 38.96 & 18.23 & 27.63 & 14.48 \\
 & I & 34.97 & 43.74 & \bestcell{5.02} & 42.73 & 31.62 & 15.72 \\
 & O & 43.54 & 36.51 & 46.59 & 7.87 & 33.63 & 15.31 \\
\midrule
ConvNeXt-V2-base & M & 24.70 & 30.02 & 49.53 & 37.60 & 35.46 & 9.33 \\
 & C & 44.18 & 13.68 & 56.89 & 47.34 & 40.52 & 16.19 \\
 & I & 49.25 & 51.55 & 13.67 & 50.80 & 41.31 & 15.98 \\
 & O & 47.81 & 51.47 & 58.68 & 33.47 & 47.86 & 9.18 \\
\midrule
ConvNeXt-V2-huge & M & 23.66 & 46.81 & 40.49 & 35.34 & 36.58 & 8.49 \\
 & C & 48.89 & 8.40 & 45.29 & 49.95 & 38.13 & 17.25 \\
 & I & 52.55 & 54.17 & 14.37 & 50.27 & 42.84 & 16.50 \\
 & O & 55.07 & 42.89 & 53.81 & 26.47 & 44.56 & 11.47 \\
\bottomrule
\end{tabular}
  }
\end{table*}

\begin{table*}[htbp]
  \centering
  \caption{Cross-dataset test ACER (\%) --- continued.}
  \label{tab:cross-dataset-acer-cont2}
  \footnotesize
  \resizebox{0.7\linewidth}{!}{%
\begin{tabular}{llcccccc}
\toprule
\multirow{2}{*}{Model} & \multirow{2}{*}{Train} & \multicolumn{6}{c}{Test} \\
\cmidrule(lr){3-8}
& & M & C & I & O & Avg. & Std. \\
\midrule
ViT-MSN-base & M & 47.85 & 58.56 & 61.63 & 55.96 & 56.00 & 5.11 \\
 & C & 54.09 & 51.70 & 64.04 & 58.04 & 56.97 & \bestcell{4.67} \\
 & I & 49.67 & 62.44 & 50.57 & 54.47 & 54.29 & 5.04 \\
 & O & 52.04 & 52.57 & 48.18 & 36.06 & 47.21 & 6.66 \\
\midrule
ViT-MSN-large & M & 49.75 & 52.44 & 46.80 & 39.01 & 47.00 & 5.03 \\
 & C & 50.91 & 51.04 & 47.34 & 39.19 & 47.12 & 4.81 \\
 & I & 50.17 & 53.39 & 42.66 & 39.58 & 46.45 & 5.56 \\
 & O & 52.82 & 46.94 & 48.81 & 37.51 & 46.52 & 5.62 \\
\midrule
ViT-MAE-base & M & 20.60 & 30.82 & 48.76 & 42.15 & 35.58 & 10.77 \\
 & C & 47.86 & 14.32 & 64.76 & 48.09 & 43.76 & 18.32 \\
 & I & 48.26 & 58.00 & 7.37 & 57.09 & 42.68 & 20.74 \\
 & O & 54.44 & 37.09 & 55.73 & 31.59 & 44.71 & 10.56 \\
\midrule
ViT-MAE-huge & M & \bestcell{16.38} & 46.95 & 14.58 & 39.71 & \bestcell{29.40} & 14.17 \\
 & C & 47.53 & 23.40 & 41.72 & 47.15 & 39.95 & 9.83 \\
 & I & 52.00 & 49.45 & \bestcell{2.07} & 60.86 & 41.10 & 22.93 \\
 & O & 38.02 & 39.85 & 34.98 & 24.81 & 34.41 & 5.81 \\
\midrule
SigLIP-base & M & 22.36 & 39.90 & 46.04 & 38.36 & 36.67 & 8.75 \\
 & C & 48.23 & 14.31 & 48.73 & 48.00 & 39.82 & 14.73 \\
 & I & 43.05 & 38.94 & 16.46 & 51.70 & 37.54 & 13.01 \\
 & O & 50.00 & 39.37 & 49.18 & 36.78 & 43.83 & 5.84 \\
\midrule
SigLIP-large & M & 27.31 & 32.16 & 50.71 & 28.64 & 34.71 & 9.41 \\
 & C & 48.39 & \bestcell{9.46} & 47.98 & 48.65 & 38.62 & 16.84 \\
 & I & 45.52 & 28.27 & 24.36 & 40.45 & 34.65 & 8.64 \\
 & O & 47.52 & 37.30 & 48.88 & \bestcell{15.90} & 37.40 & 13.20 \\
\bottomrule
\end{tabular}
  }
\end{table*}

\begin{table*}[htbp]
  \centering
  \caption{Cross-dataset test ACER (\%) --- continued.}
  \label{tab:cross-dataset-acer-cont3}
  \footnotesize
  \resizebox{0.7\linewidth}{!}{%
\begin{tabular}{llcccccc}
\toprule
\multirow{2}{*}{Model} & \multirow{2}{*}{Train} & \multicolumn{6}{c}{Test} \\
\cmidrule(lr){3-8}
& & M & C & I & O & Avg. & Std. \\
\midrule
EVA-CLIP-8B & M & 12.23 & 40.84 & 40.67 & 31.83 & 31.39 & 11.65 \\
 & C & 30.02 & 2.13 & 25.90 & 39.92 & 24.49 & 13.88 \\
 & I & 33.61 & 34.98 & 15.26 & 41.21 & 31.27 & 9.67 \\
 & O & 46.21 & 31.00 & 49.00 & 11.15 & 34.34 & 15.04 \\
\midrule
Qwen2.5-VL-3B-ViT & M & 19.61 & 35.64 & 50.17 & 33.45 & 34.72 & 10.83 \\
 & C & 42.27 & 6.51 & 43.79 & 46.47 & 34.76 & 16.38 \\
 & I & 53.02 & 44.80 & 21.33 & 48.65 & 41.95 & 12.25 \\
 & O & 41.00 & 36.64 & 42.75 & 17.15 & 34.38 & 10.20 \\
\midrule
Qwen2.5-VL-7B-ViT & M & 17.15 & 24.19 & 48.32 & 30.89 & 30.14 & 11.57 \\
 & C & 43.76 & 6.08 & 46.48 & 38.19 & 33.63 & 16.18 \\
 & I & 50.39 & 29.23 & 27.59 & 44.41 & 37.91 & 9.74 \\
 & O & 50.91 & 38.47 & 42.70 & 18.14 & 37.55 & 12.07 \\
 \midrule
InternViT-300M & M & 22.17 & 40.42 & 51.18 & 47.18 & 40.24 & 11.12 \\
 & C & 39.94 & 34.46 & 50.52 & 47.94 & 43.22 & \bestcell{6.39} \\
 & I & 50.06 & 50.83 & 16.07 & 50.19 & 41.79 & 14.85 \\
 & O & 49.92 & 41.04 & 53.69 & 22.74 & 41.85 & 11.95 \\
\midrule
InternViT-6B & M & \bestcell{4.42} & 16.64 & 45.29 & 14.65 & 20.25 & 15.18 \\
 & C & 28.07 & \bestcell{1.81} & 21.26 & 25.86 & \bestcell{19.25} & 10.36 \\
 & I & 28.54 & 25.26 & 0.06 & 50.33 & 26.05 & 17.83 \\
 & O & 48.50 & 38.89 & 40.37 & \bestcell{0.27} & 32.01 & 18.68 \\
\midrule
FSFM\_FAS & M & \bestcell{2.84} & 39.04 & 11.07 & 30.57 & 20.88 & 14.54 \\
 & C & 43.81 & \bestcell{1.65} & 52.28 & 38.97 & 34.18 & 19.38 \\
 & I & 33.21 & 49.46 & 0.17 & 54.01 & 34.21 & 21.12 \\
 & O & 39.28 & 21.97 & 33.76 & 12.04 & 26.76 & 10.55 \\
\midrule
DeepPixBiS & M & 7.89 & 47.05 & 49.65 & 36.46 & 35.26 & 16.56 \\
 & C & 33.74 & 2.01 & 30.74 & 30.20 & 24.17 & 12.87 \\
 & I & 47.36 & 61.34 & \bestcell{0.00} & 50.94 & 39.91 & 23.61 \\
 & O & 46.55 & 38.36 & 45.10 & 3.84 & 33.46 & 17.38 \\
\bottomrule
\end{tabular}
  }
\end{table*}

\FloatBarrier

